\documentclass{article}

\usepackage{amsmath}
\usepackage{fullpage}

\usepackage[numbers]{natbib}

\usepackage[utf8]{inputenc} 
\usepackage[T1]{fontenc}    
\usepackage{hyperref}       
\usepackage{url}            
\usepackage{booktabs}       
\usepackage{amsfonts}       
\usepackage{nicefrac}       
\usepackage{microtype}      
\usepackage{xcolor}         
\usepackage{xspace}
\usepackage{graphicx}
\usepackage{makecell}
\usepackage{colortbl}
\usepackage{tcolorbox}      
\usepackage{enumitem}       
\usepackage{subcaption}
\usepackage{wrapfig}
\usepackage{sidecap, caption}

\newcommand{\data}[1]{\textsc{MedMax}\xspace}
\newcommand{\llava}[1]{LLaVA\xspace}
\newcommand{\mmout}[1]{\textsc{MedMax-Instruct}\xspace}

\definecolor{backblue}{RGB}{210, 230, 250}
\newcommand{\best}{\cellcolor{backblue}}

\title{\data{}: Mixed-Modal Instruction Tuning for Training Biomedical Assistants}

\newcommand{\customfootnotetext}[2]{{%
  \renewcommand{\thefootnote}{#1}%
  \footnotetext[0]{#2}}}%

\author{
\normalsize \textbf{Hritik Bansal}
\hspace{1em}
\normalsize \textbf{Daniel Israel}$^{*}$
\hspace{1em}
\normalsize \textbf{Siyan Zhao}$^*$
\hspace{1em}
\normalsize \textbf{Shufan Li} 
\hspace{1em}
\normalsize \textbf{Tung Nguyen }
\hspace{1em}\\
\normalsize \textbf{Aditya Grover}
\\\\
\textbf{University of California Los Angeles} 
}

\begin{document}

\maketitle
\customfootnotetext{}{${*}$ Equal Contribution (alphabetical order).}

\begin{abstract}
Recent advancements in mixed-modal generative have opened new avenues for developing unified biomedical assistants capable of analyzing biomedical images, answering complex questions about them, and generating multimodal patient reports. However, existing datasets face challenges such as small sizes, limited coverage of biomedical tasks and domains, and a reliance on narrow sources. To address these gaps, we present \data{}, a large-scale multimodal biomedical instruction-tuning dataset for mixed-modal foundation models. With 1.47 million instances, \data{} encompasses a diverse range of tasks, including interleaved image-text generation, biomedical image captioning and generation, visual chat, and report understanding. These tasks span knowledge across diverse biomedical domains, including radiology and histopathology, grounded in medical papers and YouTube videos. Subsequently, we fine-tune a mixed-modal foundation model on the \data{} dataset, achieving significant performance improvements: a $26\%$ gain over the Chameleon model and an $18.3\%$ improvement over GPT-4o across 12 downstream biomedical visual question-answering tasks. Finally, we introduce a unified evaluation suite for biomedical tasks to guide the development of mixed-modal biomedical AI assistants. The data, model, and code is available at \url{https://mint-medmax.github.io/}.
\end{abstract}

\sidecaptionvpos{figure}{c}
\begin{SCfigure}[40][h]
\centering
    \includegraphics[scale=0.55]{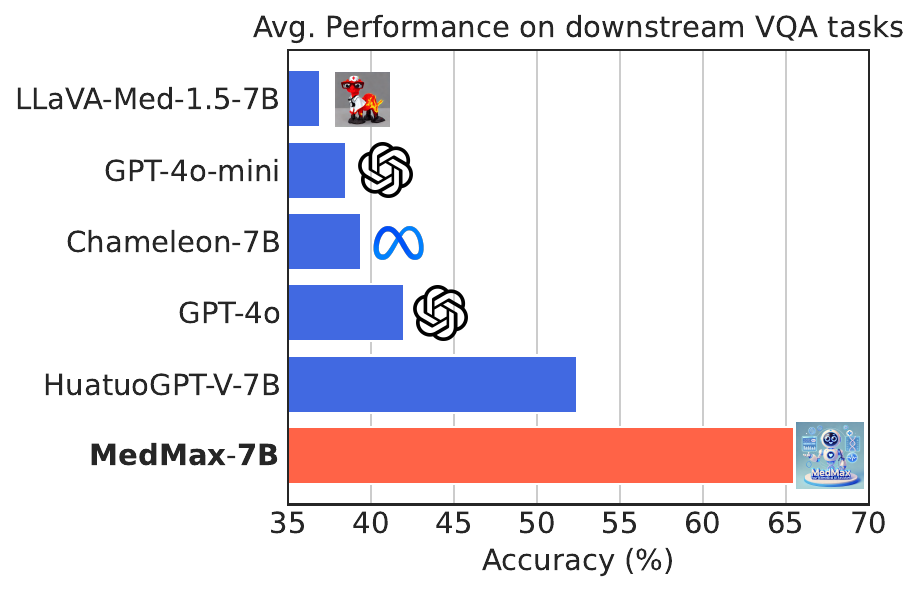}
    \caption{\textbf{Average performance of multimodal models on twelve VQA tasks.} Our \data{} instruction-tuned mixed-modal foundation model outperforms both open multimodal models (Chameleon, \llava{}-Med-v1.5, and Huatuo) and closed multimodal models (GPT-4o, GPT-4o-mini). This underscores the effectiveness of the \data{} dataset in training capable multimodal biomedical assistants.}
\label{fig:main_graph}
\end{SCfigure}

\begin{figure}[t]
    \centering
         \includegraphics[width=0.95\textwidth]{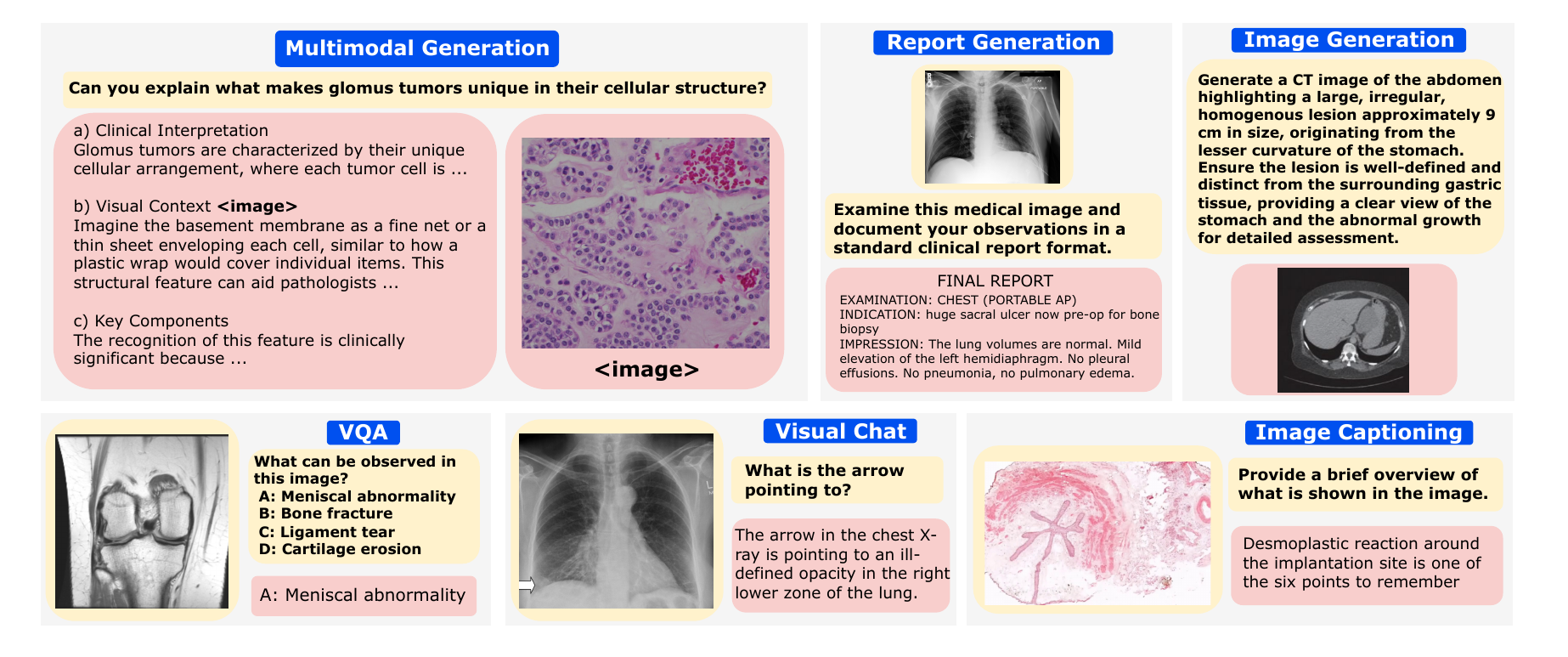}
\caption{\small{\textbf{Examples of diverse multimodal biomedical tasks covered in the \data{} dataset.} The model inputs (yellow boxes) and corresponding outputs (red boxes) illustrate various task types: multimodal generation with interleaved text and images, medical report generation, text-to-image generation, visual question answering, medical image analysis through visual chat, and image captioning task. Note that report-conditioned image generation, which falls under report understanding, is not shown here.}}
\label{fig:skills_examples}
\end{figure}

\section{Introduction}
\label{introduction}

Recently, there has been rapid advancement in the development of mixed-modal foundation models that can perceive and generate data from multiple modalities such as Chameleon \cite{team2024chameleon} and Transfusion \cite{zhou2024transfusion}.
During pretraining, these models are exposed to internet-scale data that equips them with the knowledge to perform real-world tasks involving multiple modalities within a unified architecture (e.g., image captioning and image generation). Their native multimodal capabilities have unlocked new opportunities to tackle challenging biomedical tasks, including analyzing patient scans for accurate diagnosis and generating multimodal medical reports \cite{acosta2022multimodal,wang2024perspective,gu2023biomedjourney}.

However, existing mixed-modal foundation models struggle to perform well on vision-language biomedical data due to significant distribution shifts from the more commonly occurring natural data found on the internet (e.g., everyday objects and scenes). In this context, instruction tuning \cite{wei2021finetuned, llava, lambert2024t} offers a promising approach to understanding novel user intents and unlocking new capabilities for developing advanced biomedical assistants. But, there is a lack of large-scale multimodal biomedical instruction-tuning datasets, which are crucial for enabling mixed-modal models to reason and solve complex biomedical tasks across diverse domains.

Traditional biomedical visual question answering (VQA) datasets such as VQA-RAD \cite{lau2018dataset}, SLAKE \cite{liu2021slake}, and PathVQA \cite{he2020pathvqa} are crucial for imparting domain-specific knowledge. However, they are limited in the dataset size (e.g., typically only thousands of instances). Other work like \llava{}-Med \cite{llava} collects biomedical vision-language alignment data (i.e., image-text pairs) and curated synthetic instruction-tuning datasets (i.e., image-conditioned conversational data). This aids in allowing the foundation models to assist the practitioners for novel queries about the biomedical images. However, much of the dataset consists of figures and plots rather than biomedical images, impacting its overall quality. 
Subsequently, PubMedVision \cite{zhang2023huatuogpt} curated synthetic biomedical data curation using multimodal foundation models \cite{gpt4v}. Nevertheless, the scope of their dataset is limited to the biomedical knowledge available in medical research papers while other works have shown significant biomedical knowledge is grounded in the youtube videos \cite{seyfioglu2024quiltinstruct} as well as doctor-assisted medical reports \cite{johnson2019mimic}. In addition to conversational capabilities, a native multimodal model can aid in visualizing patient reports, enabling applications such as the creation of annotated medical datasets \cite{chambon2022roentgen} and modeling disease progression \cite{gu2023biomedjourney}. However, no instruction-tuning datasets currently exist for the unified training of a multimodal biomedical assistant with diverse capabilities.

To address these challenges, we propose \data{}, a dataset designed to develop a biomedical mixed-modal foundation model. It comprises a total of $1.47$M instances spanning a wide range of biomedical tasks and domains. Specifically, \data{} includes tasks such as biomedical image captioning, image generation, visual question answering (VQA), visual chatting, report understanding, and multimodal (interleaved text-image) content generation. Moreover, the dataset encompasses diverse biomedical domains, including radiology and histopathology. A key component of \data{} is a newly curated dataset for generating interleaved image-text content (\mmout{}), which paves the way for enhanced clinical understanding and support for complex medical decision-making. Additionally, \data{} aims to equip mixed-modal models with a diverse skill set by integrating various high-quality multimodal datasets, including VQA datasets, instruction-following datasets, alignment datasets, and medical reports.

Subsequently, we fine-tune a mixed-modal foundation model, Chameleon \cite{team2024chameleon, chern2024anole}, on the \data{} dataset. The dataset comprises a total of $1.7$B multimodal discrete tokens for instruction tuning. In our experiments, the \data{}-fine-tuned model outperforms the base Chameleon and GPT-4o \cite{hurst2024gpt} by $26\%$ and $18.3\%$ percentage points, respectively, when averaged across a set of 12 downstream VQA evaluation tasks (Figure \ref{fig:main_graph}). Given the general lack of support for biomedical evaluation, we provide a comprehensive evaluation suite encompassing diverse tasks to enable unified and efficient assessments. Thus, we conduct extensive experiments on diverse tasks that mixed-modal models excel in, including biomedical image captioning, image generation, visual chatting, and multimodal generation. Overall, our work establishes a strong foundation for high-quality instruction tuning data creation, model fine-tuning, and robust evaluation of next-generation mixed-modal models.
\section{Background}
\label{sec:background}

\begin{figure*}[t]
    \centering
    \includegraphics[width=0.7\linewidth]{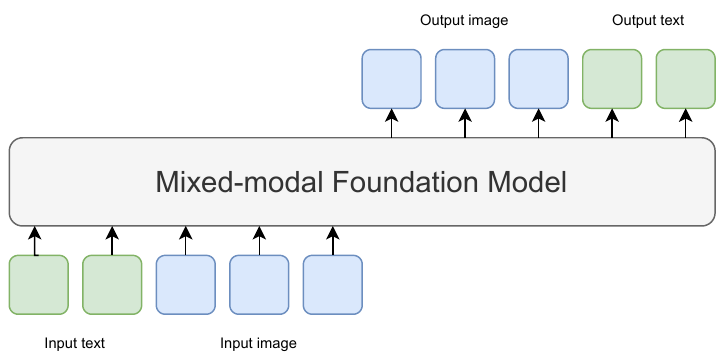}
    \caption{A mixed-modal foundation model is capable of understanding text and image inputs and can generate both textual and visual outputs through a unified architecture.}
    \label{fig:medmax_diagram}
\end{figure*}

\subsection{Mixed-Modal Foundation Models}

Mixed-modal foundation models are a class of generative models that can reason over the sequence of arbitrarily interleaved multimodal (e.g., image, text) content \cite{team2024chameleon,zhou2024transfusion}. In this work, we primarily focus on the autoregressive sequence modeling objective, as used by Chameleon \cite{team2024chameleon}, Unified-IO \cite{lu2023uio2}, and Emu-3 \cite{wang2024emu3}, for its simplicity. Formally, these methods model interleaved multi-modal tokens $x={(x_1,x_2,\dots,x_n)}$ where $n$ is the length of the sequence. Here, the text content is represented as BPE tokens, while image tokens are obtained from an image encoder. For instance, Chameleon uses 1024 tokens obtained from a VQGAN \cite{esser2021taming} encoder to represent each image. Given a dataset $\mathcal{D}$, an autoregressive pretraining objective for multi-modal data $x\in D$ can be formulated as:
\begin{equation}
    \max_\theta \mathbb{E}_{x\sim\mathcal{D}}\left[\sum_{k=1}^{n} \log P_\theta(x_k|x_{1:k-1})\right].
\end{equation}
By incorporating a diverse set of multimodal sequences at an internet scale, these pretrained models can achieve strong performance on downstream tasks such as image generation and image captioning (Figure \ref{fig:medmax_diagram}).

\subsection{Instruction Tuning}

While internet-scale pretraining equips foundation models with diverse world knowledge, further instruction tuning is conducted to develop new skills and teach them how to interact with humans as assistants \cite{wei2021finetuned,alpaca}. 
In our context, the instruction tuning data comprises paired multimodal sequences $(x, y)$, where $x$ represents the `instruction' and $y$ represents the `response.' Both $x$ and $y$ may include image tokens, text tokens, or a combination of both. For instance, in the context of visual question answering (VQA) tasks, $x$ includes an input image along with a corresponding question, while $y$ provides the correct response. The instruction tuning objective for a dataset $\mathcal{D}_{\mathcal{I}}$ can be formalized as:
\begin{equation}
    \max_\theta \mathbb{E}_{(x,y)\sim\mathcal{D}_{\mathcal{I}}}\left[\sum_{k=1}^{n} \log P(y_k|y_{1:k-1},x)\right].
\end{equation}

In particular, the instruction tuning objective computes the loss solely on the `response' portion of the sequence, thereby concentrating the learning process on the desired tasks.
\section{\data{}}
\label{sec:data}

In this work, we aim to solve diverse biomedical tasks (e.g., VQA, generation), domains (e.g., radiology, histopathology), and modalities (e.g., image and text content) through instruction tuning. Thus, we present \data{}, an instruction tuning data designed  training mixed-modal foundation models for the biomedical applications. Specifically, the dataset construction involves: (a) designing a new instruction-tuning data that allows interleaved image-text content as a model response (\S \ref{sec:mm_out_data_creation}), and curating various data sources to endow diverse skills into the model (\S \ref{sec:data_collection}).

\begin{figure}
\centering
\begin{minipage}{.42\textwidth}
  \centering
  \includegraphics[width=\linewidth]{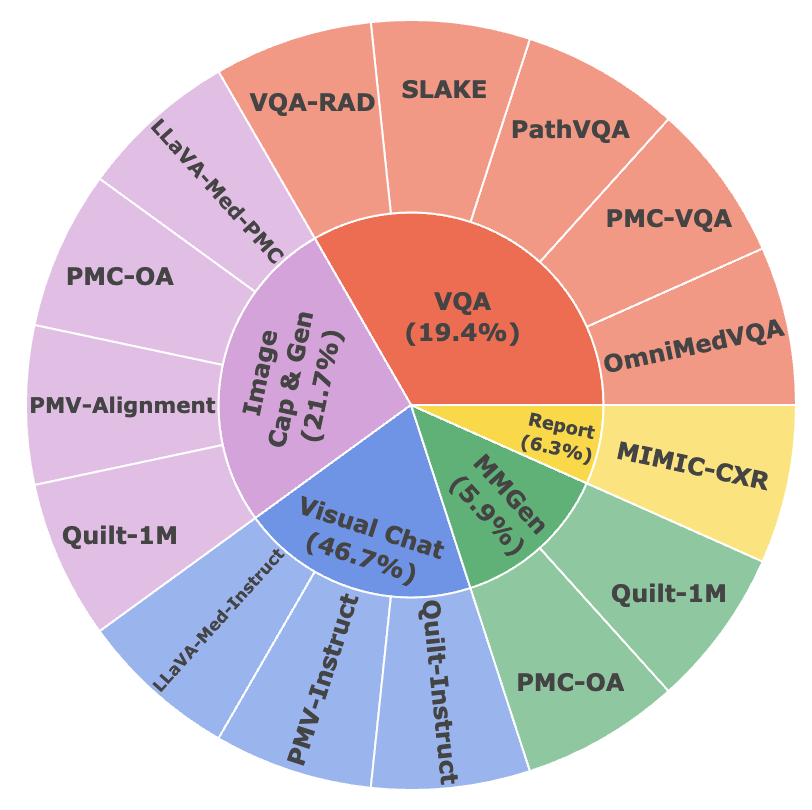}
  \captionof{figure}{We present the data-sources used to curate task-specific data in the \data{} collection.}
  \label{fig:task_data_sources}
\end{minipage}%
\hspace{2em}
\begin{minipage}{.52\textwidth}
  \centering
  \includegraphics[width=\linewidth]{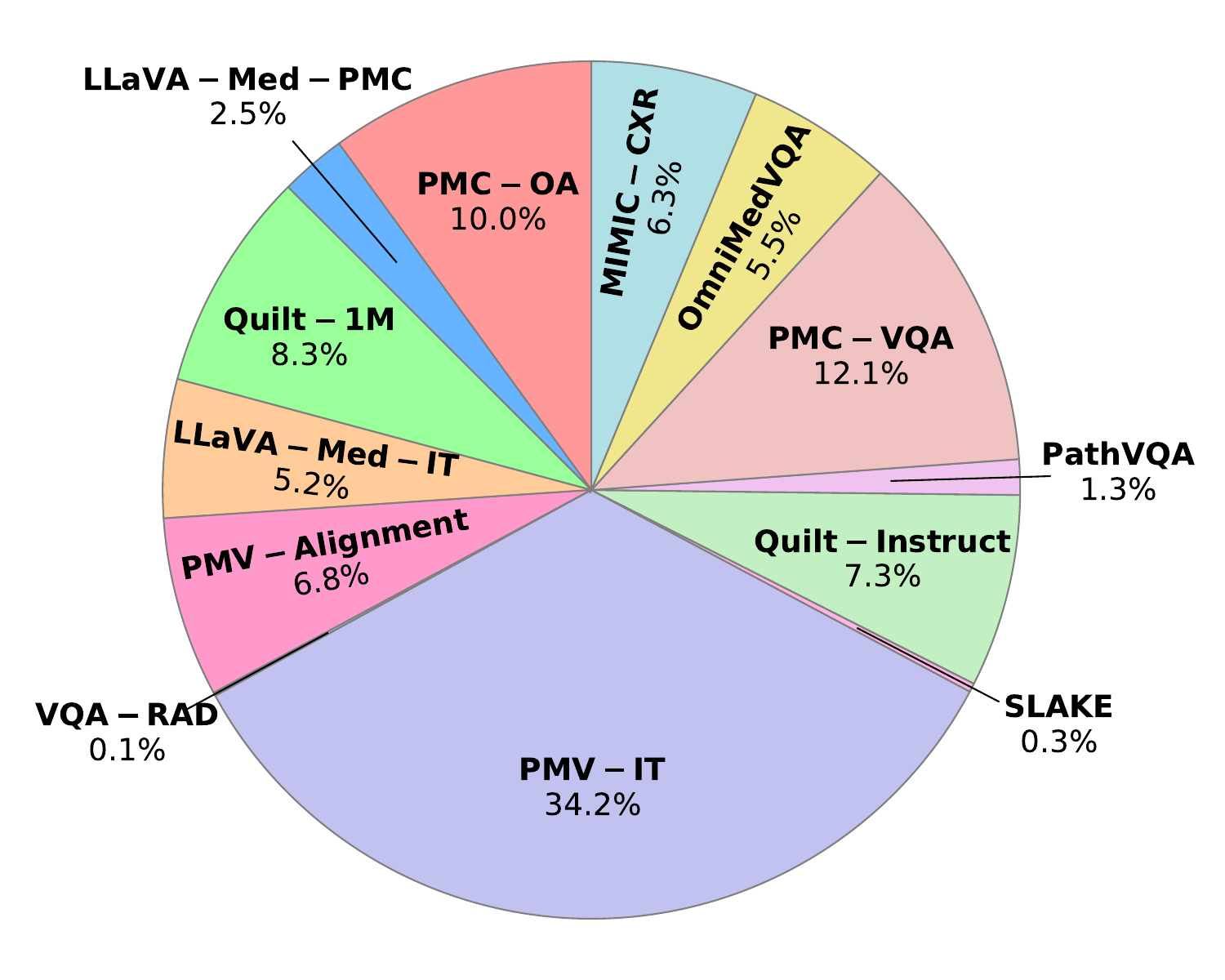}
  \captionof{figure}{We source the data from biomedical sources that cover several domains (e.g., radiology) and knowledge bases (e.g., research papers, YouTube).}
  \label{fig:pie_charts}
\end{minipage}
\end{figure}

\subsection{\mmout{}}
\label{sec:mm_out_data_creation}

With the rise of mixed-modal foundation models, generating interleaved image-text content is now possible. This capability enables novel applications like advanced diagnostics (e.g., visualizing treatment effects on biomedical markers), generating multimodal patient reports, and enhancing medical training with relevant multimodal content. However, no instruction-tuning datasets exist for training mixed-modal assistants in this area. Therefore, we introduce \mmout{}, a multimodal generation instruction-tuning dataset for biomedical mixed-modal assistants, created in three stages as described:

\paragraph{\underline{Sourcing image-text data (Stage 1)}:} Here, we intend to collect diverse biomedical image-text pairs that can be purposed for this task. In this regard, we combine image-caption instances from the PMC-OA and Quilt data. Specifically, our PMC-OA-specific curation (\S \ref{sec:data_collection}) reduces the original size of the PMC-OA (val split) from $166$K to $73$K instances.\footnote{We reserve the use of PMC-OA (train split) for image captioning and generation task, as discussed in \S \ref{sec:data_collection}.} In addition, we randomly choose $50$K instances from Quilt data. In total, we have $123$K image-text data which will be further filtered to ensure high-quality. 

\paragraph{\underline{Filter bad captions (Stage 2)}:} Since we aim to generate multimodal conversations conditioned on the captions, it is critical to ensure that they describe the visual contents of the image well. To address this, we filter the captions that GPT-4o-mini \cite{2024GPT4omini} finds to be of low quality. We present the prompt used for assessing the caption quality in Appendix Table \ref{tab:prompt_cap_quality}. Following this, we are left with $88$K instances, leading to the removal of $25\%$ captions. 

\paragraph{\underline{Caption-conditioned data generation (Stage 3):}} Here, we prompt GPT-4o to generate single-turn multiodal conversation conditioned on the captions in the remaining dataset. We choose GPT-4o as it achieves state-of-the-art performance on text-only medical datasets such as MedQA \cite{nori2023can}. We present our data generation template in Appendix Table \ref{tab:mmgen_annotation}. Specifically, we ask the GPT model to output an image placeholder (`<image>') which is then replaced with the ground-truth image from the image-caption pair. In total, we spent $\$500$ to collect GPT-4o responses using the API. Finally, we have \textbf{88K} instances consisting novel query (grounded in text) and multimodal response (interleaved text-image) for the biomedical applications. We present an example for multimodal generation in Figure \ref{fig:skills_examples}. Our data generation strategy is highly scalable, and we have limited the dataset size due to budget constraints.

\subsection{Dataset Curation for Diverse Skills}
\label{sec:data_collection}

Further, we aim to specialized biomedical datasets to teach novel skills to the mixed-modal foundation models:

\paragraph{\underline{Visual question answering (VQA)}:} 
We include biomedical visual questions that involve answering close-ended questions (yes/no), open-ended questions, and multiple-choice questions in \data{}. 
Prior work such as LLaVA-1.5 \cite{llavav15,li2024llavaonevision} have included the data from the general-purpose VQA benchmarks in their instruction tuning mix to enhance the model's visual capabilities. Hence, we combine the training sets of popular VQA datasets including VQA-RAD \cite{lau2018dataset}, SLAKE \cite{liu2021slake}, PathVQA \cite{he2020pathvqa}, and PMC-VQA \cite{zhang2023pmc}. To embed the knowledge about different anatomical regions ($20+$ regions), we split the OmnimedVQA \cite{hu2024omnimedvqa} dataset into a training (81K) and testing set (1K), and add the training set into the \data{} data. In addition to the diversity in the domains and question styles, these datasets are also a rich source of expert-annotated data (e.g., clinician-driven annotations). In total, we have \textbf{284K} VQA instances in \data{} dataset.

\textbf{\underline{Image captioning and generation}:} The ability to interpret and analyze biomedical images is crucial for accurate disease monitoring and diagnosis \cite{pavlopoulos2019survey}. Additionally, the ability to generate images is valuable for creating high-quality annotated biomedical imaging data \cite{chambon2022roentgen}.
To this end, we first curate \llava{}-Med-PMC that refers to the $600$K subset of PMC-15M \cite{zhang2023biomedclip} that was released as a part of LLaVA-Med data. On manual inspection, we found that this data is mostly composed of statistical figures, graphs and plots instead of biomedical reports (e.g., X-ray, MRI, CT). To filter this data, we utilize BioMedCLIPScore \cite{zhang2023biomedclip} score (Appendix \ref{app:llava_med_pmc_data_cleaning}). Subsequently, we are left with $37$K image-caption instances from \llava{}-Med-PMC.

Further, we utilize the PMC-OA \cite{lin2023pmc} data that also consists of image-caption data curated from PubMedCentral research papers. We provide more details about its curation in Appendix \ref{app:pmc_oa_data_curation}. Overall, the filtered training PMC-OA split contains $83$K instances. Further, we include $100$K subset of Quilt \cite{ikezogwo2024quilt} histopathology dataset collected from youtube for the biomedical image captioning and generation tasks.  
Finally, we also include the synthetically-generated PubMedVision-AlignmentVQA \cite{zhang2023huatuogpt} data which utilizes GPT-4-Vision \cite{gpt4v} to denoise and reformat noisy internet data for biomedicine. We provide more details about its curation in Appendix \ref{app:pubmedvision_data_curation}. In total, we curate \textbf{320K} instances for the image captioning ($160$K) and the image generation ($160$K).

\paragraph{\underline{Visual chat}:} 
Therefore, we aim to curate a diverse set of queries related to biomedical images that are relevant to practitioners across various biomedical domains (e.g., radiology, histopathology) and sources (e.g., research papers, internet videos). 
Concretely, we collect $76$K instances from \llava{}-Med-instruct-120K \cite{li2024llavamed} data. The remaining instances were not available publicly.  In particular, half of this data was constructed using the image description and the other half also utilized the inline mentions of the images in the PubMedCentral research papers. To further enrich our data with diverse instructions, we include PubMedVision-IT \cite{zhang2023huatuogpt} dataset which is created using GPT-4-Vision \cite{2023GPT4VisionSC}). Originally, this data consists $647$K instances, but we filter instances with multiple images in the context that left us with $504$K instances. Finally, we include $107$K conversations from the Quilt-Instruct \cite{seyfioglu2024quiltinstruct} to diversify our dataset with knowledge from the youtube videos (and transcriptions) for the histopathology domain. Overall, the \data{} consists \textbf{686K} instances for visual chat scenarios.
\paragraph{\underline{Medical report understanding}:} The ability to perform detailed inspection of a patient's imaging data requires specialized training. Thus, it is vital to expose our model to the expert-written findings (normal and abnormal anatomical cues in the image) from the patient's data. Hence, we collect chest radiographs along with the medical reports in the MIMIC-CXR \cite{johnson2019mimic} data. We provide the details about its curation in Appendix \ref{app:mimic_cxr_curation}. Finally, we have \textbf{92K} instances consisting chest radiograph-report pairs. We purpose half of the dataset for radiograph-conditioned report generation task, and the other half to generate chest radiographs conditioned on the medical report. We provide the templates in Appendix \ref{app:templates} (Table \ref{tab:image_to_report_prompts} and \ref{tab:report_to_image_prompts}).

We illustrate examples for diverse skills included in the \data{} in Figure \ref{fig:skills_examples}. In addition, we illustrate the details of the task-specific data sources in Figure \ref{fig:task_data_sources}. Further, we highlight the biomedical domains and knowledge bases covered in the \data{} in Table \ref{tab:info_data_sources}. Additionally, Figure~\ref{fig:pie_charts} provides details on the biomedical domain, database, and the proportion of individual dataset sources. Overall, we note that \data{} comprises 725K unique images and 947K unique words. Collectively, these features underscore the diversity across various quality axes, making it well-suited for instruction tuning of mixed-modal foundation models.


\section{Experimental Setup}
\label{sec:setup}

Post data creation, we instruction-tune a mixed-modal foundation model on the \data{} (\S \ref{sec:setup_model}). Then, we present the evaluation framework for robust assessment (\S \ref{sec:evaluation}).

\subsection{\data{} Mixed-Modal Model}
\label{sec:setup_model}

In our work, we instruction-tune Anole-7B \cite{chern2024anole}, an instantiation of the Chameleon-7B \cite{team2024chameleon} mixed-modal foundation model that can natively understand and generate multimodal content. 
We use LoRA \cite{hu2021lora} for parameter-efficient finetuning of the model. Further, we finetune the base model for $3$ epochs on \data{}. We provide more details in Appendix \ref{app:finetuning_details}.\footnote{While various mixed-modal models, such as Transfusion \cite{zhou2024transfusion} and Monoformer \cite{zhao2024monoformer}, offer different approaches, we select Chameleon (or Anole) for its simplicity. Thus, we note that \data{} can enable finetuning of any of these mixed-modal models too.}

\begin{table*}[t]
\caption{\textbf{Lists of tasks in the \data{} evaluation suite.} We perform comprehensive evaluation of the \data{}-finetuned mixed-modal model across diverse biomedical multimodal tasks. We note that BioMedCLIP can be used to assess the similarity between two images, which is referred to as Image-Image BioMedCLIPScore. We abbreviate visual question answering as VQA, multi-choice questions as MCQ, exact matching as EM, and large language model as LLM.}
\label{tab:eval_summary}
\centering
\resizebox{\textwidth}{!}{%
\begin{tabular}{lcc}
\toprule
\textbf{Task}               & \textbf{Source}    & \textbf{Metric}                        \\\hline
\textit{Biomedical Visual Question Answering} &&\\
VQA (Closed) & VQA-RAD \cite{lau2018dataset}     & Accuracy (EM)        \\
VQA (Closed) & SLAKE \cite{liu2021slake}      & Accuracy (EM)        \\
VQA (Closed) & PathVQA \cite{he2020pathvqa}    & Accuracy (EM)        \\
VQA (Closed) & Quilt-VQA \cite{seyfioglu2024quiltinstruct}  & Accuracy (EM)        \\
VQA (Open)   & VQA-RAD \cite{lau2018dataset}    & Accuracy (LLM)                \\
VQA (Open)   & SLAKE \cite{liu2021slake} & Accuracy (LLM)                \\
VQA (Open)   & PathVQA \cite{he2020pathvqa}   & Accuracy (LLM)                \\
VQA (Open)   & Quilt-VQA  \cite{seyfioglu2024quiltinstruct}                   & Accuracy (LLM)                \\
VQA (MCQ)            & PMC-VQA \cite{zhang2023pmc}     & Accuracy (EM)        \\
VQA (MCQ) & OmniMedVQA \cite{hu2024omnimedvqa}                     & Accuracy (EM)        \\
VQA (MCQ)                & PathMMU \cite{sun2025pathmmu} & Accuracy (EM)        \\
VQA (MCQ)                & ProbMed \cite{yan2024worse}    & Accuracy (EM)        \\\hline
\textit{Biomedical Image Captioning and Generation}&&\\
Image captioning      & PMC-OA \cite{lin2023pmc}     & BioMedCLIPScore               \\
Image generation   & PMC-OA \cite{lin2023pmc}    & BioMedCLIPScore               \\
Image captioning      & Quilt\cite{ikezogwo2024quilt}       & BioMedCLIPScore               \\
Image generation   & Quilt \cite{ikezogwo2024quilt}     & BioMedCLIPScore               \\
Image captioning      & MIMIC-CXR \cite{johnson2019mimic}  & BioMedCLIPScore               \\
Image generation   & MIMIC-CXR \cite{johnson2019mimic} & BioMedCLIPScore               \\\hline
\textit{Biomedical Visual Chatbot}    & LLaVA-Med \cite{li2024llavamed} & LLM score                     \\\hline
\textit{Biomedical Multimodal Generation (\textbf{NEW})}     & \makecell{PMC-OA\cite{lin2023pmc}\\Quilt \cite{ikezogwo2024quilt}}               & \makecell{LLM score\\  \small{Image-Image BioMedCLIPScore}}\\
\bottomrule
\end{tabular}%
}
\end{table*}

\subsection{Evaluation}
\label{sec:evaluation}

While there are several multimodal biomedical datasets, there is a general lack of a comprehensive evaluation benchmark. To address this, we introduce a thorough evaluation suite to assess the capabilities of native multimodal models across various tasks and domains.

\paragraph{\underline{Biomedical VQA}:} 
We include the test set of VQA-RAD (radiology), SLAKE (semantic knowledge over radiology), PathVQA (pathology), and the entire QuiltVQA (histopathology) dataset. These datasets ask closed-ended (yes/no) and open-ended questions that require one word, phrase or sentence answer. Here, we use exact match to assess the accuracy of the models on the closed-ended questions. However, the evaluation on the open-ended questions is intrinsically harder due to the subjectivity of the answers. To avoid such challenges, we utilize an LLM (GPT-4o-mini) that compares the predicted answer against the ground-truth answer to decide where the model outputs are reliable or not. For each open-ended question, it gives a score of 0 or 1. We provide the evaluation template in Appendix Table \ref{tab:open_ended_llm_score}. 
Additionally, we also include medical VQA with multiple-choice questions datasets such as test set of the PMC-VQA (diverse biomedical domains), validation set of PathMMU (pathology) \cite{sun2025pathmmu}, and ProbMed (radiology) \cite{yan2024worse} dataset. In addition, we assess the performance of an hidden split of $1000$ questions from the OmniMedVQA \cite{hu2024omnimedvqa} dataset. Overall, we perform evaluations on twelve VQA tasks across diverse biomedical domains, skills, and question formats. 

\paragraph{\underline{Biomedical image captioning and generation}:} We compare the ability of the \data{} model and the base model to caption as well as generate biomedical images for diverse domains. In total, we collect $1200$ instances from the testing split of PMC-OA (400), MIMIC-CXR (400), and unseen split of Quilt-1M (400) datasets. In particular, half of the dataset will be used for captioning evaluation and the other half will be used for generation evaluation. Similar to \citet{chambon2022roentgen}, we extract the summary of the findings (impressions) from the report data and treat them as the ground-truth captions for the associated chest radiographs from MIMIC-CXR. This will highlight the medical report understanding and report-conditioned image generation capability of our model. Subsequently, we compute the BioMedCLIPScore to assess the closeness between the input image (caption) and predicted caption (image). We present the details for model inference in Appendix \ref{app:eval_gen_templates}.

\textbf{\underline{Biomedical Visual Chatbot}:} We use the visual chatbot evaluation from \llava{}-Med consisting $193$ novel questions about $50$ unseen biomedical images. Specifically, the questions belong to two category: conversation and detailed description of the images.  Subsequently, an LLM scores the predicted answer and the GPT-4 written reference answer out of 10 conditioned on the question, image caption and additional image context. Finally, we compute the average relative prediction score as the ratio of the score for predicted answer and score for the reference answer. 

\textbf{\underline{Biomedical Multimodal Generation}:} We utilize $500$ hidden instances of the \mmout{} data for model evaluation. For a given text query, we prompt the base (or finetuned) model to generate interleaved response. Subsequently, we compare the text content in the predicted multimodal response with the reference text response using LLM (same as visual chat). In addition, we compare the generated image with the reference image using the image-image similarity score from the BioMedCLIP model. We present the evaluation templates and inference details in Appendix \ref{app:eval_gen_templates}. Further, we conduct contamination analysis and find that there are no exact matches between image-text pairs across these datasets. We provide the summary of the tasks and metrics in Table \ref{tab:eval_summary}.
\section{Experiments}
\label{sec:experiments}
\begin{table*}[t]
\caption{\small{\textbf{Performance of the \data{} model and baselines on the downstream VQA tasks.} We find that the \data{} mixed-modal model outperforms closed as well as open multimodal models on twelve VQA datasets. This highlights that the model can generalize well to unseen instances and tasks ranging across biomedical domains.}}
\label{tab:results_vqa_table}
\centering
\resizebox{\textwidth}{!}{%
\begin{tabular}{lcccccc}
\toprule
\textbf{Model}           & \makecell{\textbf{Chameleon}\\\textbf{(7B)}} & \makecell{\textbf{\llava{}-Med}\\\textbf{(v1.5-7B)}} & \makecell{\textbf{GPT-4o}\\\textbf{(mini)}} & \textbf{GPT-4o} & \makecell{\textbf{HuatuoGPT}\\\textbf{(Vision-7B)}} & \makecell{\textbf{\data{}}\\\textbf{(7B)}}  \\\hline
Average         & 39.4        & 36.6          & 38.5       & 42.0  & 52.4               & \best{65.5} \\\hline
VQA-RAD (Closed) \cite{lau2018dataset}  & 48.6         & 61.0           & 55.8        & 54.2   & 74.5                & \best{75.3}  \\
SLAKE (Closed) \cite{liu2021slake} & 59.1         & 48.7           & 50.4        & 50.1   & 70.7                & \best{88.4}  \\
PathVQA (Closed) \cite{he2020pathvqa} & 58.9         & 62.7 & 48.7        & 59.2   & 65.9                & \best{91.8}  \\
QuiltVQA (Closed) \cite{seyfioglu2024quiltinstruct}& \best{71.4}         & 63.0          & 38.5       & 44.6 & 55.7                & 61.2  \\
VQA-RAD (Open)  \cite{lau2018dataset}   & 32.0         & 23.0           &   13.0          &   17.6     & 19.0                & \best{46.5}  \\
SLAKE (Open) \cite{liu2021slake}  & 5.3          & 25.1           &  49.3           &  63.7      & 53.3                & \best{82.2}  \\
PathVQA (Open)  \cite{he2020pathvqa}  & 18.0         & 6.2            &    7.3         &  9.1      & 6.0                 & \best{40.6}  \\
QuiltVQA (Open)\cite{seyfioglu2024quiltinstruct}   & 15.3         & 17.2           &  28.0           &     \best{36.1}   & 22.2                & 26.0 \\
PMC-VQA     \cite{zhang2023pmc}    & 31.0           & 18.9           & 39.6        & 40.8   & \best{51.6}                & 49    \\
OmniMedVQA  \cite{hu2024omnimedvqa} & 45.7         & 28.7           & 45.1        & 40.9   & 75.6                & \best{99.5}  \\
PathMMU \cite{sun2025pathmmu}    & 34.5         & 29.8           & 35.6        & 39.1  & \best{55.4}                & 49.3  \\
ProbMed\cite{yan2024worse}        & 52.8         & 58.5           & 50.6        & 48.3   & \best{78.7}                & 75.8  \\
\bottomrule
\end{tabular}%
}
\end{table*}

\subsection{Main results}
\label{sec:main_results}

\paragraph{Biomedical Visual Question Answering.}
\label{sec:results_biomed_vqa}

We compare the performance of the finetuned \data{} model against several vision-language models on the battery of the VQA datasets in our evaluation suite. Specifically, the baselines includes the open models: the mixed-modal Chameleon-7B \cite{team2024chameleon}, and multimodal input but text-only biomedical specific models: \llava{}-Med-v1.5 \cite{llava}, and HuatuoGPT-Vision-7B \cite{zhang2023huatuogpt}. In addition, we consider two closed vision-language models: GPT-4o-mini and GPT-4o. We present our results in the Table \ref{tab:results_vqa_table}. Our experiments reveal that \data{} model outperforms the base model (Chameleon) by $26.1\%$ percentage points on the average VQA performance across twelve tasks. This highlights that \data{} is suitable for building biomedical specialized model from the base model via instruction-tuning. Moreover, we observe that it achieves the \textit{best} average performance across diverse baselines. In particular, it beats the closed vision model, GPT-4o, by $18.3\%$ percentage points. It indicates that the \data{} model is the most capable multimodal model for biomedical VQA across open as well as closed models. Further, we highlight the task-specific performance on individual VQA tasks too.

In addition, we observe that \data{} achieves the highest accuracy amongst the baselines for 7 out of 12 tasks. Interestingly, we find that Chameleon-7B achieves the highest QuiltVQA (Closed) but a low QuiltVQA (Open) accuracy which suggests that closed-ended questions are easier to solve for the base model. Additionally, \data{} achieves $99.5\%$ performance on the unseen split of OmniMedVQA, demonstrating that this dataset can be solved with high accuracy when exposed to in-distribution examples. The model demonstrates the ability to answer questions about diverse anatomical regions after being exposed to some in-distribution data from OmniMedVQA. Furthermore, HuatuoGPT-Vision-7B outperforms other baselines on 3 out of 12 datasets. This can be attributed to its high-quality fine-tuning dataset and the Qwen-2 backbone \cite{yang2024qwen2}, pre-trained on 7 trillion text tokens, which enhances the model's biomedical knowledge and its ability to interpret diverse formatting styles.

Additionally, we observe that the \data{} model achieves performance improvements of $10.7\%$, $15\%$, and $23\%$ on the unseen QuiltVQA (Open), PathMMU, and ProbMed datasets, respectively. This underscores the \data{} model's ability to generalize effectively to novel downstream tasks through exposure to expert knowledge, question formats, and biomedical domains via instruction-tuning. Furthermore, we find that the \data{} is competitive to the task-specific finetuning of the \llava{}-Med on VQA-RAD, SLAKE and PathVQA datasets (Appendix \ref{app:task_specific_ft_comparison}). This indicates that the practitioners can utilize an unified \data{} model instead of hosting separate task-specific \llava{}-Med models. Overall, these results highlight at the high-quality of the \data{} dataset for finetuning the mixed-modal foundation models for biomedical assistants.

\paragraph{Biomedical Image captioning and generation.}
\label{sec:results_biomed_cap_gen}

\begin{figure}[t]
    \centering
    \begin{subfigure}[h]{0.31\textwidth}
        \centering
        \includegraphics[width=\textwidth]{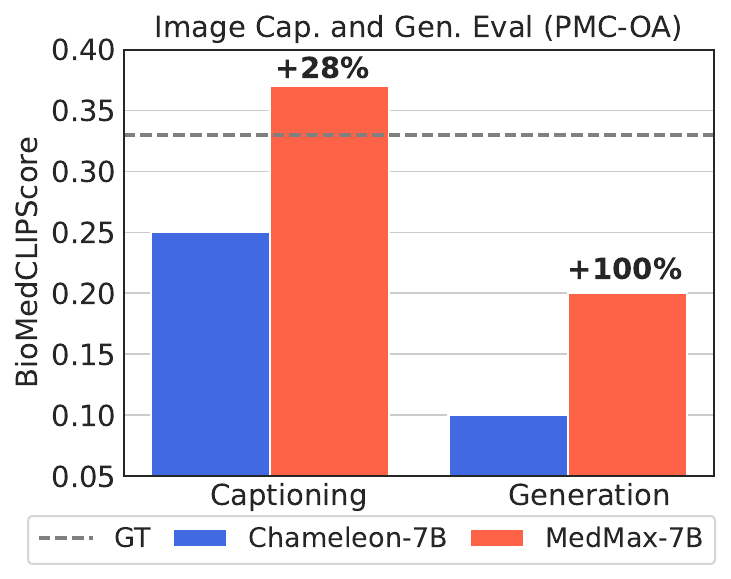}
        \caption{Performance on PMC-OA.}
        \label{fig:cap_gen_results_a}
    \end{subfigure}%
    \hfill
    \begin{subfigure}[h]{0.31\textwidth}
        \centering
        \includegraphics[width=\textwidth]{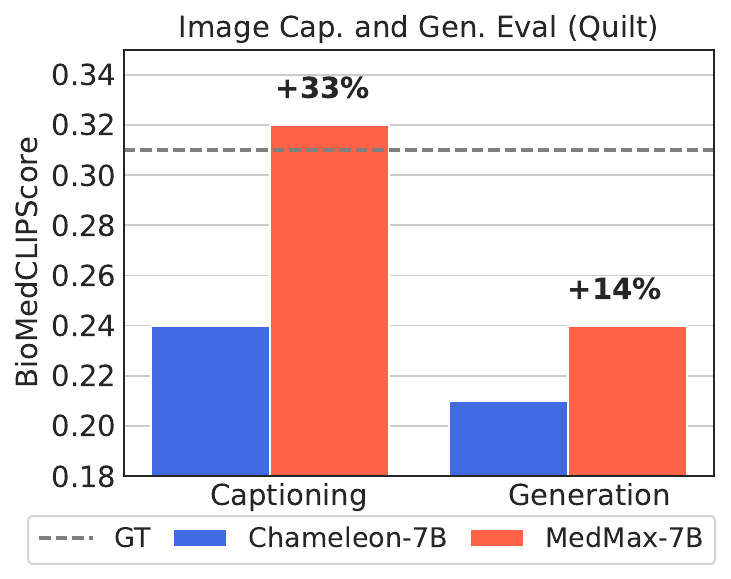}
        \caption{Performance on Quilt.}
        \label{fig:cap_gen_results_b}
    \end{subfigure}
    \hfill
    \begin{subfigure}[h]{0.31\textwidth}
        \centering
        \includegraphics[width=\textwidth]{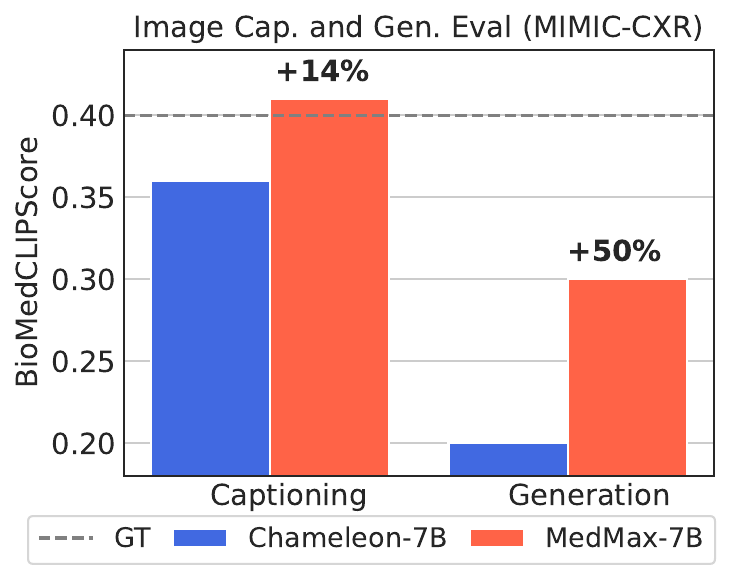}
        \caption{Performance on CXR.}
        \label{fig:cap_gen_results_c}
    \end{subfigure}
    \caption{\textbf{Performance on the image captioning and image generation tasks.} We find that \data{} model consistency outperforms the base Chameleon mixed-modal model across diverse biomedical domains.}
    \label{fig:cap_gen_results}
\end{figure}

Here, we compare the capability of the \data{} model and base model to interpret and generate biomedical image data across different biomedical domains. We present the results in Figure \ref{fig:cap_gen_results}. Our empirical findings suggest that the \data{} consistently outperforms the base model on the biomedical image captioning as well as the image generation task. In particular, the \data{} outperforms Chameleon by achieving a relative gain of $28\%$, $33\%$ and $14\%$ on biomedical captioning for the images in the PMC-OA, Quilt, and MIMIC-CXR datasets, respectively. Additionally, the \data{} also outperforms the Chameleon by achieving a relative gain of $100\%$, $14\%$, and $50\%$ on the image generation tasks across PMC-OA, Quilt,and MIMIC-CXR captions, respectively. Further, we compare the captioning abilities of our model against other models in Appendix \ref{app:add_image_captioning_results}. Overall, our results demonstrate that the \data{} model excels at reasoning about images and generating realistic biomedical visuals aligned with user intent.

\paragraph{Biomedical Multimodal Generation.}
\label{sec:results_biomed_mm_gen}

\begin{figure}[t]
    \centering
    \begin{subfigure}[h]{0.48\textwidth}
        \centering
        \includegraphics[width=0.8\textwidth]{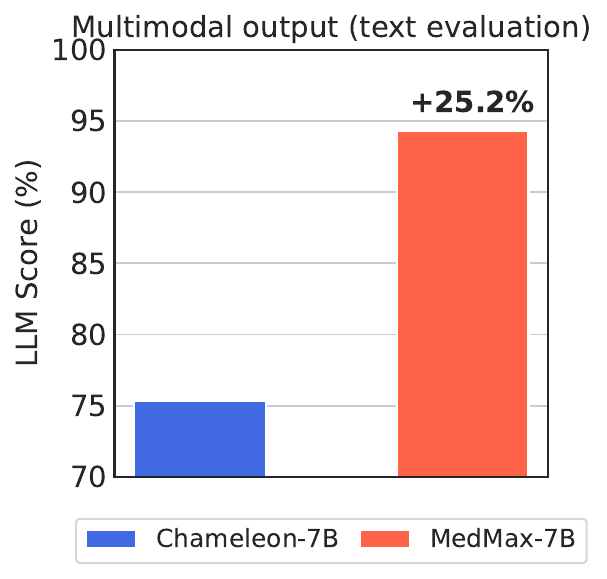}
        \caption{Evaluating text content.}
        \label{fig:mm_out_results_a}
    \end{subfigure}%
    \begin{subfigure}[h]{0.48\textwidth}
        \centering
        \includegraphics[width=0.8\textwidth]{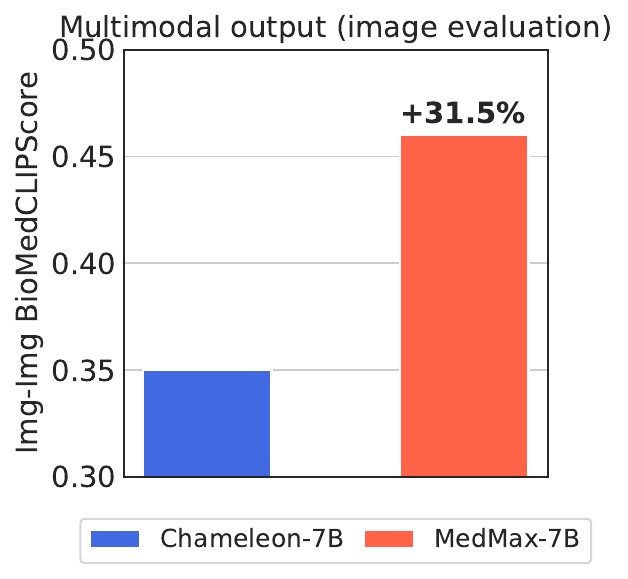}
        \caption{Evaluating image content.}
        \label{fig:mm_out_results_b}
    \end{subfigure}
    \caption{\textbf{Performance on the multimodal generation task.} Comparison between the performance of the \data{} and Chameleon mixed-modal model on the multimodal generation task. We find that \data{} finetuning improves the multimodal content generation capabilities for the biomedical domain.}
    \label{fig:mm_out_results}
\end{figure}

Here, we study the capability of the \data{} model to generate multimodal (interleaved image-text) content (Figure \ref{fig:mm_out_results}). Our empirical findings highlight that the \data{} outperforms Chameleon by achieving a relative improvement of $25.2\%$ on the quality of the text content in the multimodal output, as measured by LLM score. Furthermore, we observe that the \data{} outperforms Chameleon by achieving a relative improvement of $31.5\%$ on the synthesized biomedical image in the multimodal output, as measured by the image-image similarity score using BioMedCLIP. This indicates that instruction tuning with \data{} data equips a mixed-modal model with strong multimodal generation capabilities in the biomedical domain. Overall, our results lay the foundation for future exploration in unlocking and evaluating multimodal generation capabilities via instruction-tuning. 

\begin{figure}[h]
\centering
\begin{minipage}{.45\textwidth}
  \centering
  \includegraphics[width=\linewidth]{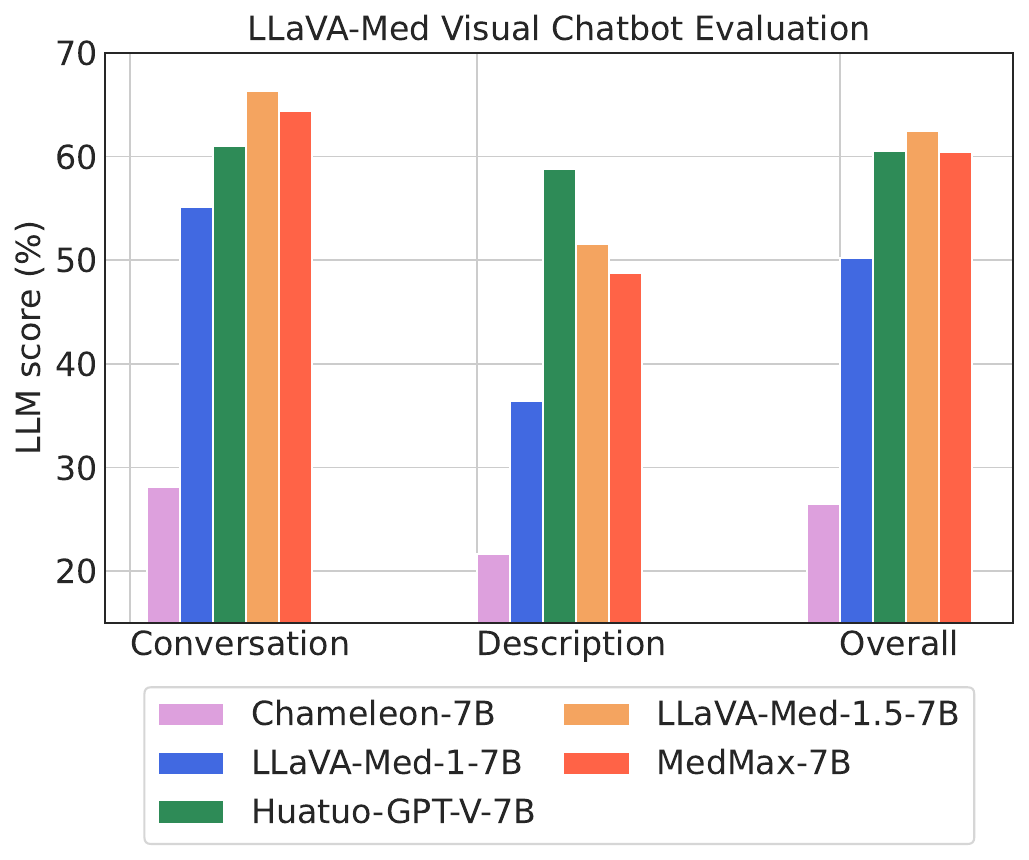}
  \captionof{figure}{\textbf{Performance on the visual chat task.} We find that the chatting capabilities of our model is quite competitive, suggesting its ability to answer novel queries about biomedical images.}
  \label{fig:visual_chat_eval}
\end{minipage}%
\hspace{2em}
\begin{minipage}{.4\textwidth}
  \centering
  \includegraphics[width=\linewidth]{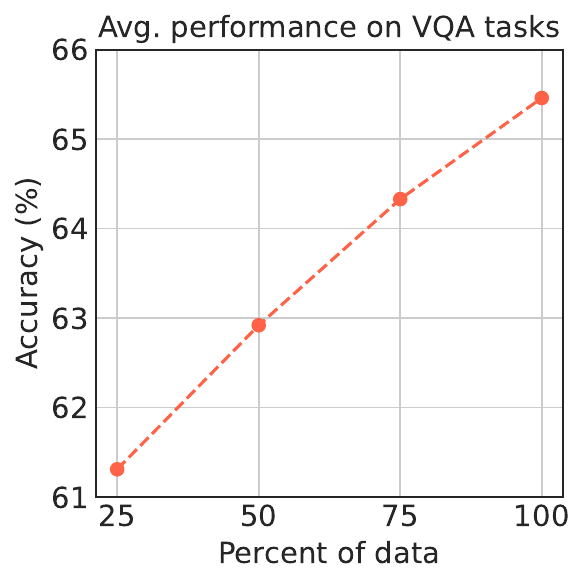}
  \captionof{figure}{\textbf{Model performance with data scale.} We find that \data{} is a high-quality dataset that enhances VQA performance as it scales.}
  \label{fig:data_scaling}
\end{minipage}
\end{figure}

\paragraph{Biomedical Visual Chat.}
\label{sec:results_biomed_visual_chat}

We study the ability of the \data{} model and the baselines to answer novel queries about the biomedical images. We present the results in Figure \ref{fig:visual_chat_eval}. We observe that the overall LLM score of \data{} model is higher than the base Chameleon model and \llava{}-med-v1 by $34\%$ and $10.2\%$ percentage points. This indicates instruction-tuning enables strong visual chatting capability to the mixed-modal models. In addition, the \data{} model is quite competitive in comparison to the best-performing model, \llava{}-Med-v1.5, with a difference of $2.1\%$ percentage points.
We observe that \llava{}-Med-v1.5 excels in the conversation split, while HuatuoGPT-Vision performs better on the description split of the dataset. These improvements are likely to stem from their high-quality language backbones, Mistral-v0.2-Instruct and Qwen-2, which enhance query interpretation and text generation. Future advancements in mixed-modal models will further improve biomedical visual chat performance with the \data{}.

\subsection{Qualitative Examples}
\label{sec:qualitative_examples}

We present a few qualitative examples for \data{}-7B generations in Figure \ref{fig:positive_qualitative_examples} and \ref{fig:negative_qualitative_examples}. The good qualitative examples highlight the model's strong understanding of biomedical images and its ability to generate detailed answers across various tasks, especially in producing multimodal reports in response to questions. However, the examples also demonstrate the model's limitations, including its occasional misinterpretation of image details and generation of images with meaningless textual artifacts.

\section{Ablation Studies}
\label{sec:ablation}

In this section, our goal is to the study the role of different factors that can impact the downstream performance during mixed-modal instruction tuning. 

\subsection{Data scaling}
\label{sec:exp_data_scaling}

Now, we explore how the benefits of mixed-modal instruction tuning scale with the size of the dataset. Specifically, we finetune the Anole-7B (instantiation of Chameleon-7B) with three subsets of the \data{} including $25\%$, $50\%$, and $75\%$ of the data. Subsequently, we evaluate the average performance on the twelve VQA tasks for these subsets and the entire dataset. We present the performance vs the dataset scale in Figure \ref{fig:data_scaling}. Our empirical finding suggests that the downstream performance of the instruction-tuned mixed-modal model monotonically increases with the size of the data. This highlights that the \data{} dataset is of high quality, and further scaling has the potential to yield greater improvements on downstream tasks. 

\subsection{Impact of Specific Data Subsets}
\label{sec:ablation_data_subsets}

\begin{figure*}[h]
    \centering
    \begin{subfigure}[h]{0.42\textwidth}
        \centering
        \includegraphics[width=\textwidth]{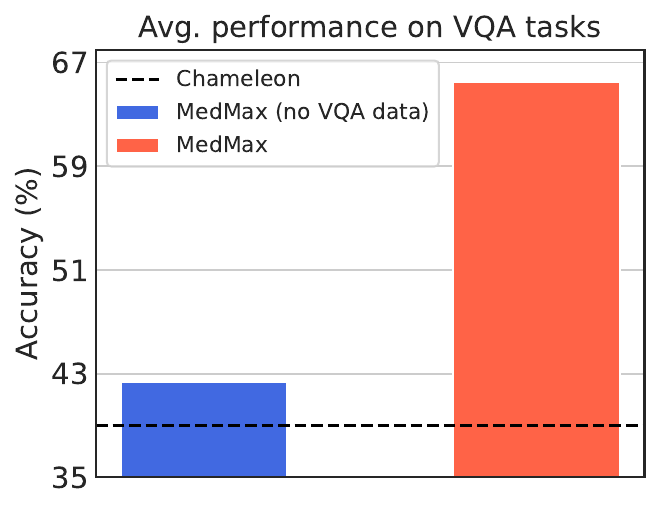}
        \caption{Data ablation for VQA task.}
        \label{fig:data_subset_ablation_a}
    \end{subfigure}%
    \hspace{3em}
    \begin{subfigure}[h]{0.42\textwidth}
        \centering
        \includegraphics[width=\textwidth]{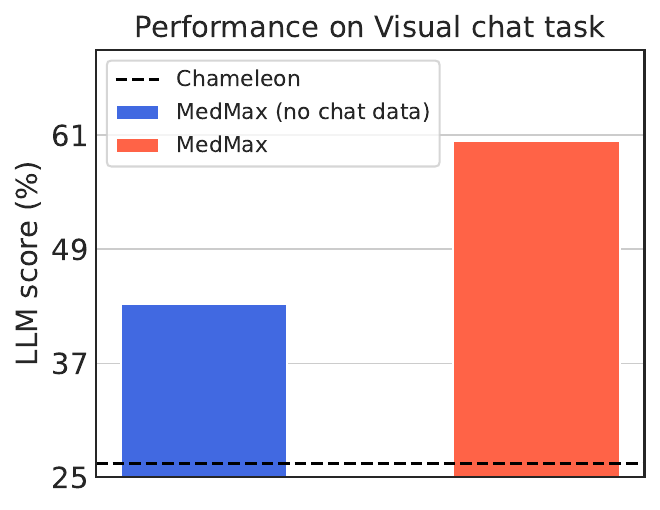}
        \caption{Data ablation for visual chat task.}
        \label{fig:data_subset_ablation_b}
    \end{subfigure}
    \caption{\textbf{Results for the data ablation study.} Finetuning the mixed-modal model with an ablated version of the \data{} data where the (a) VQA task instances and (b) visual chat instances are removed. The results highlight the usefulness of task-specific data in the mixture for downstream performance.}
    \label{fig:data_subset_ablation}
\end{figure*}

Since \data{} is a collection of several tasks, we aim to understand the impact of specific tasks on downstream performance. To this end, we create two subsets of the \data{} dataset: (a) all tasks except VQA, and (b) all tasks except visual chat. We select VQA tasks because our model achieves very high performance on them, and visual chat tasks because they constitute the largest proportion of the dataset. Subsequently, we fine-tune the base model on these two ablated versions of \data{}. The results are presented in Figure \ref{fig:data_subset_ablation}.

In Figure \ref{fig:data_subset_ablation_a}, we observe that the model fine-tuned on the VQA-ablated dataset suffers a performance drop of $23\%$ compared to the original dataset on VQA tasks. This highlights the importance of including high-quality VQA tasks in the \data{} mixture. Similarly, Figure \ref{fig:data_subset_ablation_b} shows that the model fine-tuned on the chat-ablated dataset experiences a performance drop of $17\%$ on visual chat tasks. This underscores the critical role of visual chat data in achieving strong performance on visual chat tasks. Overall, our experiments demonstrate the value of incorporating diverse tasks in the data mixture to enable generalization across downstream biomedical tasks.

\subsection{Impact of Specialized Visual Encoder}
\label{sec:ablation_visual_encoder}

\sidecaptionvpos{figure}{c}
\begin{SCfigure}[40][t]
\centering
    \includegraphics[scale=0.6]{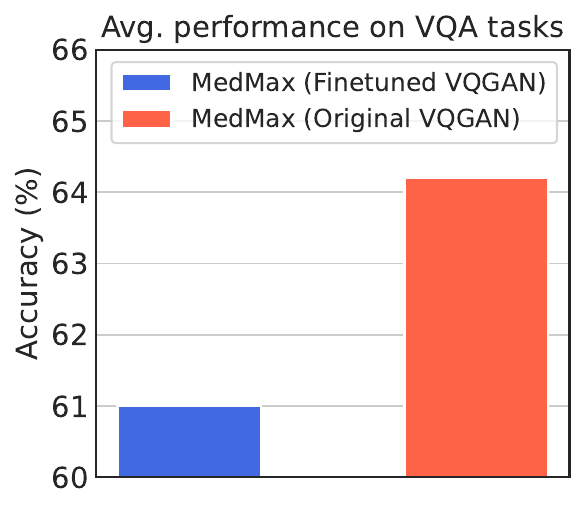}
    \caption{\textbf{Effect of finetuned visual encoder on downstream performance.} We find that the finetuning the Chameleon model with the original visual tokens gives better downstream performance than finetuned visual token on biomedical images. This highlights that the model does not like the distribution shift in the visual token distribution.}
\label{fig:finetuned_vqgan_ablation}
\end{SCfigure}

In the visual instruction tuning literature \cite{li2024llavamed}, a two-stage fine-tuning approach is commonly employed. In the first stage, the visual encoder is specialized to match the distribution of images in the instruction-tuning dataset, while keeping the transformer backbone of the architecture frozen. In the same spirit, we explore whether finetuning of the Chameleon's VQGAN encoder with biomedical images before instruction-tuning with \data{} leads to a better downstream model. In this regard, we finetune the base VQGAN with $300$K images from the \data{} dataset for $8$ epochs.\footnote{We used the original VQGAN codebase for finetuning: \url{https://github.com/CompVis/taming-transformers}.} We find that the L1 reconstruction loss for the finetuned VQGAN was $7.8$ in comparison to the base VQGAN $8.1$ on a set of held-out biomedical images. This indicates that the fine-tuned encoder effectively represents the new domain better than the base visual encoder. Next, we select a random subset of 800K samples from the \data{} dataset and tokenize the images using the newly fine-tuned visual encoder. We then fine-tune the multimodal model using both the original subset and the re-tokenized subset under identical settings.

We present the performance of the two instruction-tuned mixed-modal models in Figure \ref{fig:finetuned_vqgan_ablation}. We find that the model finetuned with the new discrete visual tokens achieves an inferior performance to the model finetuned with the original (base) visual tokens from the VQGAN. Specifically, we observe a gap of $3\%$ averaged across the VQA datasets in our evaluation suite. This can be attributed to the distribution shift in the discrete visual tokens relative to the base model, where the base VQGAN visual tokens are better aligned with the base model's representations. Further exploration of fine-tuned, specialized visual encoders for discrete multimodal models is left for future work.

\section{Related Work}

\paragraph{Multimodal biomedical assistants.} 

While early biomedical language models like ChatDoctor~\citep{li2023chatdoctor}, MedicalGPT~\citep{MedicalGPT}, and HuatuoGPT~\citep{zhang2023huatuogpt} advanced text-only medical reasoning (often built upon large language models such as LLaMA or Alpaca variants), they lacked multimodal capabilities for integrating visual information. This limitation prompted the development of multimodal biomedical models, ranging from encoder-only architectures like BiomedCLIP~\citep{zhang2023biomedclip} to generative vision-language models capable of producing medical explanations. For instance, Med-Flamingo~\citep{moor2023medflamingo} extended OpenFlamingo~\citep{open-flamingo} to a few-shot medical VQA paradigm via continued pre-training on curated image-text pairs. MedVInT~\citep{zhang2023pmc}, based on a pre-trained vision-language model, leveraged the PMC-VQA dataset to improve generative VQA. LLaVA-Med~\citep{li2024llavamed}, built upon a LLaVA model~\citep{llava}, refined these capabilities using filtered PubMed data and GPT-4 generated instructions. RadFM~\citep{wu2023generalist} broadened image modalities to 2D and 3D radiology data. HuatuoGPT-Vision~\citep{chen2024huatuogptvisioninjectingmedicalvisual} adapts the LLaVA-v1.5~\citep{llavav15} architecture with Qwen2-7B~\citep{qwen} backbone and employs PubMedVision for large-scale medical VQA. Med-Gemini~\citep{team2023gemini} integrated advanced multimodal and retrieval mechanisms on top of a Gemini model to enhance long-context and medical image understanding. More recently, MedTrinity-25M~\citep{xie2024medtrinity25mlargescalemultimodaldataset} proposed a benchmark of over 25 million image-ROI-description triplets that will be useful for pretraining. While \data{} has a smaller scale instruction-tuning data, it prioritizes efficient instruction-tuning through careful curation. While most existing approaches center on evaluation tasks like VQA or text-based medical chats, \data{} pushes beyond the boundary by demonstrating mixed-modal generation and interleaving text-image content to further enrich clinical comprehension.

\paragraph{Multimodal instruction tuning.}

Multimodal model training typically begins by aligning modalities in a shared embedding space and then perform instruction tuning to enhance conversational capabilities~\citep{caffagni2024revolution}. LLaVA~\citep{llava} was among the first to utilize multimodal instruction-following data, generated by GPT-4, to enable rich visual conversations. MiniGPT-4~\citep{minigpt4} constructed instruction sets by combining image-text datasets from Conceptual Caption~\citep{sharma-etal-2018-conceptual}, SBU~\citep{NIPS2011_5dd9db5esbu}, and LAION~\citep{laion} with handwritten instruction templates, while InstructBLIP~\citep{instructblip} incorporated VQA datasets to enhance visual reasoning. Multi-Instruct~\citep{xu2023multiinstruct} further diversified the instruction set by incorporating 47 multimodal tasks.
Beyond single images, MIMIC-IT~\citep{li2023mimicitmultimodalincontextinstruction}, LAMM~\citep{yin2023lamm}, and Macaw-LLM~\citep{lyu2023macaw} introduced 3D, audio, and video scenarios for broader multimodal understanding. More recent datasets, such as LLaVAR~\citep{llavar}, augmented visual instruction tuning with OCR results and expanded capabilities to handle text-rich images. High-quality instruction tuning data can be combined from multiple sources: LLaVA-1.5~\citep{llavav15} improved upon LLaVA~\citep{llava} by incorporating diverse academic instruction tuning data, while LLaVA-OneVision~\citep{li2024llavaonevision} extended this approach by combining data across single-image, multi-image, and video scenarios. In this work, \data{} integrates multiple medical image datasets to create high-quality instruction tuning data that enables mixed-modal generation capabilities.

\paragraph{Mixed-modal foundation models.} Mixed-Modal foundational models use a single neural network to process inputs of multiple modalities. The training objectives of such models comes in different flavors. Earliest works such as BEIT-3\cite{wang2022image} make use of masked data modeling or contrastive learning objective from self-supervised learning field. More recent works uses generative modeling objectives instead. Among these, some work such as UniDiffuser \cite{bao2023one} use a diffusion objective to learn a joint distribution of image and text in latent space, and Transfusion~\citep{zhou2024transfusion} combines diffusion for images with autoregressive modeling for text. Alternatively, models such as Unified-IO \cite{lu2023uio2}, Chamaleon \cite{team2024chameleon}, Emu3 \cite{wang2024emu3}, CM3Leon~\citep{yu2023scaling}, and Anole~\citep{chern2024anole} formulate multi-modal learning as a general sequence modeling problem over multi-modal tokens.  This autoregressive discrete decoding approaches facilitates generation of interleaved text-image sequences while maintaining architectural simplicity. In this work, we leverage this architectural simplicity of autoregressive mixed-modal models to effectively train \data{}, enabling comprehensive biomedical instruction tuning across diverse tasks and modalities.

\section{Conclusion}


We present \data{}, the first instruction-tuning dataset enabling interleaved content generation for biomedical AI through \mmout{}. \data{} also supports biomedical VQA, dialog, captioning, generation, and report understanding. Our experiments demonstrate that \data{}-tuned models achieve strong performance across multiple tasks, laying a strong foundation for the next generation of multimodal biomedical AI assistants. We emphasize that the \data{} model is a research prototype intended to foster community involvement in developing capable biomedical assistants. In medical emergencies, users should always consult a healthcare professional rather than relying solely on the model's output. With continuous community feedback and the collection of high-quality expert data, we aim to enhance the model's accuracy and safety over time.

\section*{Acknowledgements}
AG would like to acknowledge an AI2050 Fellowship from Schmidt Sciences, NSF Career Award \#2341040, and Amazon Research Award. HB is supported in part by AFOSR MURI grant FA9550-22-1-0380. SZ is supported in part by Amazon Fellowship. We would also like to thank Yidou Weng, Ethan Israel, Helen Cai, and Mohamed Soufi for their assistance in the qualitative assessment of our model.

\bibliographystyle{plain}
\bibliography{ref}
\clearpage

\appendix

\begin{table*}[h]
\caption{\small{\textbf{Additional information about diverse biomedical dataset sources.} We highlight that \data{} consists data across several biomedical domains and knowledge bases.}}
\label{tab:info_data_sources}
\centering
\begin{tabular}{lcc}
\toprule
\textbf{Data source}            & \textbf{Domain}         & \textbf{Knowledge Base}          \\\hline
LLaVA-Med-PMC          & Diverse        & PubMed Central             \\
PMC-OA                 & Diverse        & PubMed Central         \\
Quilt-1M               & Histopathology & YouTube   \\
LLaVA-Med-IT           & Diverse        & PubMed Central     \\
PubMedVision-Alignment & Diverse        & PubMed Central      \\
PubMedVision-IT        & Diverse        & PubMed Central      \\
Quilt-Instruct         & Histopathology & YouTube      \\
VQA-RAD                & Radiology      & MedPix     \cite{nihMedPix}         \\
SLAKE                  & Radiology      & \makecell{MSD \cite{antonelli2022medical} \\ CXR-8 \cite{wang2017chestx} \\ Chaos \cite{kavur2021chaos}}      \\
PathVQA                & Pathology      & PEIR Digital Library \cite{jones2001peir}   \\
PMC-VQA                & Radiology      & PubMed Central   \cite{nihPubMedCentral}   \\
OmniMedVQA             & Diverse        & Diverse               \\
MIMIC-CXR              & Chest X-ray    & MIMIC-CXR   \cite{johnson2019mimic}         \\\bottomrule
\end{tabular}
\end{table*}

\section{More Details on Data Curation}
\label{app:curation}

\subsection{LLaVA-Med-PMC}
\label{app:llava_med_pmc_data_cleaning}


We curated a dataset of 37.8K medical images filtered from an initial pool of 538K images, sourced from the data released by LLaVA-Med~\citep{li2024llavamed}, which originates from PMC-15M~\citep{zhang2023biomedclip}. The initial dataset contained a significant number of statistical figures, as the images were extracted from research articles. To filter these out and retain only the desired medical images, we utilized the pretrained BiomedCLIP~\citep{zhang2023biomedclip} model to classify images based on the taxonomy defined in PMC-15M.

Our focus was on retaining images from the classes “Magnetic Resonance,” “CT,” “X-Ray,” “ECG,” “Light Microscopy,” “Dermatology,” and “Endoscopy,” which represent the primary topics used in LLaVA-Med. To determine class-specific confidence thresholds, we manually labeled 80 images and evaluated the model's predictive confidence using ROC curves, identifying optimal thresholds through Youden's J statistic to balance sensitivity and specificity. Additionally, we applied a heuristic to exclude images with high prediction scores for statistical figures within the top five predictions. This filtering process resulted in a high-quality dataset of 37.8K images, focused on key medical imaging modalities.

\subsection{PMC-OA}
\label{app:pmc_oa_data_curation}

Further, we utilize the PMC-OA data that also consists of image-caption data curated from PubMedCentral research papers. Its notable features includes its accessibility (open-access), size ($1$M+ instances) and diversity of biomedical imaging data (e.g., ultrasound, fMRI, endoscope, PET). To maintain one-to-one correspondence in the data, we filter instances where a caption was aligned with multiple sub-images. Further, we filter small biomedical images that were less than 200 pixels in width or height in this data.

\subsection{PubMedVision}
\label{app:pubmedvision_data_curation}

In many cases, the real image-caption has inherent data noise and formatting issues. Hence, we include the synthetically-generated PubMedVision-AlignmentVQA \cite{zhang2023huatuogpt} data which utilizes GPT-4-Vision \cite{gpt4v} to denoise and reformat noisy internet data for biomedicine.
In particular, we filter the original data ($647$K) to remove multiple image instances in this data to get $504$K instances, and randomly select a subset of $100$K instances for the \data{} mix. While this dataset can be utilized for image captioning (input image and question as context, and the answer as context), it could not be utilized for image generation directly. To address this, we prompt GPT-4o-mini \cite{hurst2024gpt} to convert the image descriptions in the PubMedVision-AlignmentVQA data into image generation prompts: `\textit{Convert the image description into an image generation prompt with AI}'. We show an example in Appendix Table \ref{tab:example_pubmedvision_alignment_image_gen}.

\subsection{MIMIC-CXR}
\label{app:mimic_cxr_curation}

We collect chest radiographs along with the medical reports in the MIMIC-CXR \cite{johnson2019mimic} data. Originally, the dataset consisted of 377K instances. We filtered it to exclude reports discussing more than one image, reducing the dataset to 102K instances. Subsequently, we subsampled the data to decrease the proportion of `No findings' reports \cite{chambon2022roentgen} from $20\%$ to $10\%$.  

\section{Limitations}

In this work, we focus on diverse multimodal biomedical skills, including VQA, multimodal generation, visual chat, image captioning, image generation, and report understanding. While these skills cover a broad spectrum of tasks, there remain additional possibilities that could further enhance biomedical applications. For instance, we do not address setups that involve understanding and generating multiple images, which are critical for applications such as counterfactual biomedical image generation \cite{gu2023biomedjourney} and reasoning from multiple images \cite{yue2023mmmu}. Achieving this capability presents significant challenges, including the lack of openly available, large-scale, high-quality multi-image biomedical datasets and the limited context length of the pretrained (base) Chameleon model. To bridge this gap, more efficient methods for representing image data are required, rather than always encoding all images as $1024$ tokens, which occupy a substantial portion of the model's context length. We leave these explorations for future work.

\section{\mmout{} Data Curation Prompts}

We present the GPT prompt to filter the bad captions from the real image-caption data in Table \ref{tab:prompt_cap_quality}. Further, we provide the prompt for generating multimodal generation conversation using GPT in Table \ref{tab:mmgen_annotation}.

\begin{table*}[h]\centering
\caption{\textbf{Prompt to assess the quality of the caption aligned with a biomedical image in the real image-caption data.}}
\begin{minipage}{0.99\textwidth}
\vspace{-2mm}
\begin{tcolorbox}[
    colback=brown!10,     
    colframe=brown!50,
]
    \small
Evaluate whether an image description provides substantive information by analyzing it against the following criteria:

1. Specificity: Does it contain precise details rather than vague descriptions? \\
2. Context: Does it provide relevant background or situational information? \\
3. Technical Details: Are any specific measurements, conditions, or technical terms included? \\
4. Purpose: Would the information be useful for professional analysis, decision-making, or documentation? \\

For medical descriptions specifically, consider: \\
- Anatomical details \\
- Condition characteristics \\
- Observable features \\
- Diagnostic relevance \\

Format your response as follows: \\
1. Analysis: Briefly explain why the description is or isn't informative (2-3 sentences) \\
2. Conclusion: End with either "The answer is: Yes" or "The answer is: No" \\ 

Example: \\\\
Description: Juvenile polyp or retention polyp is present. \\
Output: The description identifies a specific medical condition (juvenile/retention polyp) and confirms its presence, which is diagnostically relevant. While brief, this information is clinically useful for medical assessment and treatment planning.  \\The answer is: Yes \\ \\
Note: Evaluate only the information provided in the description without making assumptions about missing details.\\\\
\textbf{Description: [CAPTION]}
\end{tcolorbox}
\vspace{-3mm}
\label{tab:prompt_cap_quality}
\end{minipage}
\end{table*}

\begin{table*}[h]\centering
\caption{\textbf{Prompt to generate multimodal generation conversation using the caption in the real image-caption data.}}
\begin{minipage}{0.99\textwidth}
\vspace{-2mm}
\begin{tcolorbox}[
    colback=green!10,     
    colframe=green!50,
]
    \small
\#\# Task\\
Create natural, single-turn conversations that demonstrate how users might seek help understanding biomedical data descriptions without access to actual images. \\ 

\#\# Input Format\\
A brief, clinical description of biomedical data, focusing on: Measurements, Observed structures, Technical parameters, and Relevant findings. \\

\#\# Output Requirements\\ 

\#\#\# 1. User Question\\
Generate a natural question that: Addresses specific medical findings or terminology, Avoids references to images, figures, or descriptions, Reflects a general understanding level, and Focuses on understanding clinical significance \\   

Avoid phrases like: "In this image...", "Based on the description...", "According to the figure...", and "Can you explain what I'm seeing..."\\ 

Use formats like: \"What does it mean when the common bile duct is 15mm?", "Can you explain what MRCP tells us about the pancreas?", "Is a dilated pancreatic duct concerning?" \\ 

\#\#\# 2. AI Response\\
Structure the response with: \\ 

\#\#\#\# a) Clinical Interpretation: Begin with a clear, direct answer, Define medical terminology, Explain normal vs. abnormal values, Discuss clinical implications, Use accessible language while maintaining accuracy. \\\\

\#\#\#\# b) Visual Context: Insert `<image>' placeholder where relevant, Reference anatomical relationships, and Provide size comparisons to familiar objects when applicable. \\\\

\#\#\#\# c) Key Components: Diagnostic significance, Related conditions, Normal reference ranges, and Potential next steps or considerations \\ 

\#\#\# Style Guidelines \\ 

\#\#\#\# Language: Professional but accessible, Define technical terms, Use analogies when helpful, and Maintain clinical accuracy \\ 

\#\#\#\# Tone: Informative, Objective, and Reassuring without minimizing concerns \\ 

\#\#\#\# Structure: Clear topic sentences, Logical flow, Concise paragraphs, and Supporting details \\ 

\textbf{Input: [CAPTION] }
\end{tcolorbox}
\vspace{-3mm}
\label{tab:mmgen_annotation}
\end{minipage}
\end{table*}

\begin{table*}[h]\centering
\caption{\textbf{Converting the image description in the PubMedVision-AlignmentVQA to the image generation prompt using LLM.}}
\begin{minipage}{0.99\textwidth}
\vspace{-2mm}
\begin{tcolorbox}[
    colback=cyan!10,     
    colframe=cyan!50,
]
    \small
\begin{itemize}[leftmargin=0.5mm]
\setlength{\itemsep}{2pt}
\item \textbf{Image description in the PubMedVision-AlignmentVQA:} The image shows a chest radiograph in the anteroposterior (AP) view. The heart, mediastinal structures, and trachea appear to be displaced to the contralateral side, indicating dextrocardia, a condition where the heart is situated on the right side of the chest rather than the normal left side. The lung fields appear relatively clear, with no obvious abnormalities visible.
\item \textbf{Image generation prompt using LLM:} Create an image of a chest radiograph in the anteroposterior (AP) view. Display the heart, mediastinal structures, and trachea displaced to the opposite side, illustrating the condition of dextrocardia, where the heart is located on the right side of the chest. Ensure the lung fields appear relatively clear and show no obvious abnormalities.
\end{itemize}
\end{tcolorbox}
\vspace{-3mm}
\label{tab:example_pubmedvision_alignment_image_gen}
\end{minipage}
\end{table*}

\section{Data creation templates for captioning, generation, and report understanding}
\label{app:templates}

Table \ref{tab:concise_description_prompts} and Table \ref{tab:detailed_description_prompts} present complementary approaches to image captioning, with the former focusing on concise, brief descriptions and the latter encouraging comprehensive, detailed analyses of image content. Table \ref{tab:image_gen_prompts} demonstrates various prompts for generating images from text descriptions, using diverse language to ensure accurate visual representation of textual input. Table \ref{tab:image_to_report_prompts} showcases prompts for medical report generation from diagnostic images, emphasizing structured radiological reporting formats and professional clinical observations. Table \ref{tab:report_to_image_prompts} illustrates prompts for generating medical images from clinical reports, focusing on accurate visualization of documented pathological findings and diagnostic features.
\begin{table*}[h]\centering
\caption{\textbf{List of prompts examples for concise image captioning.}}
\begin{minipage}{0.99\textwidth}\vspace{0mm}
\begin{tcolorbox}[
    colback=yellow!10,     
    colframe=yellow!50,
]
    \centering
    \small
    \hspace{-6mm}
\begin{itemize}[leftmargin=0.5mm]
\setlength{\itemsep}{2pt}
\item Describe the image concisely: [IMAGE]
\item Provide a brief description of the given image: [IMAGE]
\item Offer a succinct explanation of the picture presented: [IMAGE]
\item Summarize the visual content of the image: [IMAGE]
\item Give a short and clear explanation of the subsequent image: [IMAGE]
\item Share a concise interpretation of the image provided: [IMAGE]
\item Present a compact description of the photo's key features: [IMAGE]
\item Relay a brief, clear account of the picture shown: [IMAGE]
\item Render a clear and concise summary of the photo: [IMAGE]
\item Write a terse but informative summary of the picture: [IMAGE]
\end{itemize}
\end{tcolorbox}
    
\vspace{-2mm}

\label{tab:concise_description_prompts}
\end{minipage}
\end{table*}

\begin{table*}[h]\centering
\caption{\textbf{List of prompts examples for detailed image captioning.}}
\begin{minipage}{0.99\textwidth}\vspace{0mm}
\begin{tcolorbox}[
    colback=yellow!10,     
    colframe=yellow!50,
]
    \centering
    \small
    \hspace{-6mm}
\begin{itemize}[leftmargin=0.5mm]
\setlength{\itemsep}{2pt}
\item Describe the following image in detail: [IMAGE]
\item Provide a detailed description of the given image: [IMAGE]
\item Give an elaborate explanation of the image you see: [IMAGE]
\item Share a comprehensive rundown of the presented image: [IMAGE]
\item Offer a thorough analysis of the image: [IMAGE]
\item Explain the various aspects of the image before you: [IMAGE]
\item Clarify the contents of the displayed image with great detail: [IMAGE]
\item Characterize the image using a well-detailed description: [IMAGE]
\item Break down the elements of the image in a detailed manner: [IMAGE]
\item Walk through the important details of the image: [IMAGE]
\end{itemize}
\end{tcolorbox}
    
\vspace{-2mm}

\label{tab:detailed_description_prompts}
\end{minipage}
\end{table*}

\begin{table*}[h]\centering
\caption{\textbf{List of prompts examples for image generation from text descriptions.}}
\begin{minipage}{0.99\textwidth}\vspace{0mm}
\begin{tcolorbox}[
    colback=violet!10,     
    colframe=violet!50,
]
    \centering
    \small
    \hspace{-6mm}
\begin{itemize}[leftmargin=0.5mm]
\setlength{\itemsep}{2pt}
\item Generate a visual representation based on the following description: [CAPTION]
\item Create a depiction that accurately illustrates this description: [CAPTION]
\item Generate an accurate representation aligned with this description: [CAPTION]
\item Create a detailed depiction that reflects the information in this description: [CAPTION]
\item Produce a clear visual based on the provided description: [CAPTION]
\item Design a representation that captures the essence of the following text: [CAPTION]
\item Generate a graphic aligned with this description: [CAPTION]
\item Create an image that visualizes the details in the following text: [CAPTION]
\item Develop a visual based on the description provided: [CAPTION]
\item Illustrate the scenario described in the following text: [CAPTION]
\end{itemize}
\end{tcolorbox}
    
\vspace{-2mm}

\label{tab:image_gen_prompts}
\end{minipage}
\end{table*}
\begin{table*}[h]\centering

\caption{\textbf{List of prompts for image-to-report generation.}}
\begin{minipage}{0.99\textwidth}\vspace{0mm}
\begin{tcolorbox}[
    colback=violet!10,     
    colframe=violet!50,
]
    \centering
    \small
    \hspace{-6mm}
\begin{itemize}[leftmargin=0.5mm]
\setlength{\itemsep}{2pt}
\item Generate a detailed medical report for this image following standard radiological reporting format.
\item As a radiologist, provide a comprehensive medical report for this diagnostic image.
\item Write a structured medical report describing all findings visible in this image.
\item Examine this medical image and document your observations in a standard clinical report format.
\item Create a detailed clinical report based on your analysis of this diagnostic image.
\item Review this medical image and generate a complete radiological report including all relevant findings.
\item Analyze this diagnostic image and provide a structured medical report with your observations.
\item Acting as an experienced radiologist, document your interpretation of this image in a medical report.
\item Evaluate this medical image and create a comprehensive clinical report detailing all findings.
\item Provide a thorough radiological report based on your examination of this diagnostic image.
\end{itemize}
\end{tcolorbox}
    
\vspace{-2mm}

\label{tab:image_to_report_prompts}
\end{minipage}
\end{table*}

\begin{table*}[h]\centering
\caption{\textbf{List of prompts for report-to-image generation.}}
\begin{minipage}{0.99\textwidth}\vspace{0mm}
\begin{tcolorbox}
    \centering
    \small
    \hspace{-6mm}
\begin{itemize}[leftmargin=0.5mm]
\setlength{\itemsep}{2pt}
\item Generate a medical image that accurately represents all findings described in this report.
\item Create a diagnostic image that visualizes all the clinical observations mentioned in this report.
\item Synthesize a medical image that corresponds to the findings detailed in this radiological report.
\item Based on this clinical report, generate a medical image showing all described features and abnormalities.
\item Produce a diagnostic image that illustrates all the medical findings documented in this report.
\item Create a medical image that faithfully represents the pathological findings described in this report.
\item Generate a diagnostic image that matches all the clinical observations in this medical report.
\item Visualize this medical report as a diagnostic image showing all mentioned findings and characteristics.
\item Transform this radiological report into a corresponding medical image with all described features.
\item Based on the clinical descriptions in this report, generate an accurate medical image representation.
\end{itemize}
\end{tcolorbox}
    
\vspace{-2mm}

\label{tab:report_to_image_prompts}
\end{minipage}
\end{table*}

\section{Additional Evaluation Details}

\subsection{VQA Open-Ended Evaluation with LLM}
\label{app:vqa_open_llm}

Table \ref{tab:open_ended_llm_score} presents the template for evaluating the models on the open-ended questions of the VQA datasets. Our prompt is motivated from the GPT evaluation prompt in \url{https://github.com/jinlHe/PeFoMed/tree/main}.

\begin{table*}[h]\centering
\caption{\textbf{Template for evaluating the correctness of the predicted answer in comparison to the ground-truth answer for the open-ended questions in the VQA datasets.}}
\begin{minipage}{0.99\textwidth}
\vspace{-2mm}
\begin{tcolorbox}[
    colback=orange!10,     
    colframe=orange!50,
]
    \small
    Given a question about an medical image, there is a correct answer to the question and an answer to be determined. If the answer to be determined matches the correct answer or is a good enough answer to the question, output 1; otherwise output 0. Evaluate the answer to be determined (1 or 0). \\\\

    Question:\\
    question about the medical image: \textbf{[question]}\\\\

    Answers:\\
    correct answer (ground truth): \textbf{[true answer]}\\
    answer to be determined: \textbf{[generated answer]}\\\\

    Task:\\
    Given a question about an medical image, there is a correct answer to the question and an answer to be determined. If the answer to be determined matches the correct answer or is a good enough answer to the question, output 1; otherwise output 0. Evaluate the answer to be determined (1 or 0).\\

    Output Format:\\
    Correctness: \textbf{[your judgment]}
\end{tcolorbox}
\vspace{-3mm}
\label{tab:open_ended_llm_score}
\end{minipage}
\end{table*}

\subsection{Evaluation Templates and Generation Modes}
\label{app:eval_gen_templates}

We present the templates and generation modes for diverse tasks in our evaluation suite in Table \ref{tab:temp_and_gen_modes}. Following the approach used in Chameleon, we suppress the probability of visual tokens in the output to zero, ensuring that only text content is generated for VQA tasks. Additionally, `image-gen' indicates that the probabilities for the text tokens are suppressed to zero to ensure that the model just generates an image in the response. Further, `any-gen' highlights that the model is free to generate multimodal content in the response. We perform greedy decoding in our experiments. Across our experiments, we use greedy decoding (temperature = 0) to generate text content and set the temperature to 0.7 for generating image content in the responses.

\begin{table*}[h]
\caption{\textbf{Template and generation modes for the downstream evaluation of the \data{} model.}}
\label{tab:temp_and_gen_modes}
\centering
\resizebox{\linewidth}{!}{%
\begin{tabular}{ccc}
\toprule
\textbf{Task}                  & \textbf{Template}                             & \textbf{Generation Mode}      \\\hline

\makecell{VQA-RAD (Open/Closed) \\PathVQA (Open/Closed)\\SLAKE (Open/Closed)\\ProbMed} & \textit{<image>[question]} & Text \\\hline
QuiltVQA (Closed)     &\makecell{\textit{<image>Answer the question based on this image and respond 'yes' or 'no'.}\\\textit{[question]}}  & Text\\\hline
QuiltVQA (Open)       & \makecell{\textit{<image>Answer the question based on this image.}\\\textit{[question]}}                         & Text  \\\hline
\makecell{PMC-VQA\\OmniMedVQA\\PathMMU}               & \makecell{\textit{<image>}\textit{[question]}\\\textit{[choice A]}\\\textit{[choice B]}\\\textit{[choice C]}\\\textit{[choice D]}} & Text\\\hline

Captioning            & \textit{<image>Please describe this picture.} & Text  \\\hline
Generation            & \textit{<caption>}                            & Image \\\hline
Multimodal generation & \textit{<question>}                           & Any  \\\hline
Visual chat (Conversation)            & \textit{<image>[question]}                            & Text \\\hline
Visual chat (Description)            & \textit{<image>Analyze the image in a comprehensive and detailed manner.
}                            & Text \\
\bottomrule
\end{tabular}%
}
\end{table*}

\begin{table*}[h]
\centering
\caption{\textbf{Comparison between \data{} model and the task-specific finetuned \llava-Med models on the closed-ended questions of VQA-RAD, SLAKE, and PathVQA datasets.}}
\label{tab:task_specific_ft_comparison}
\begin{tabular}{lccc}
\toprule
                        & \textbf{VQA-RAD (\%)} & \textbf{SLAKE (\%)} & \textbf{PathVQA (\%)} \\\hline
LLaVA-Med-Finetuned (3 epochs) \cite{li2024llavamed}  & 66.5   & 64.2  & 89.5    \\
\data{} (Ours)                & 75.3 (\textcolor{blue}{+8.8})  & \textbf{88.4} (\textcolor{blue}{+24.2})  & \textbf{91.8} (\textcolor{blue}{+2.3})    \\
LLaVA-Med-Finetuned (15 epochs) \cite{li2024llavamed} &\textbf{ 84.2}  (\textcolor{blue}{+17.7})  & 85.3 (\textcolor{blue}{+21.1})  & 91.2 (\textcolor{blue}{+1.7})    \\
\bottomrule
\end{tabular}
\end{table*}

\section{Additional VQA Results}
\label{app:task_specific_ft_comparison}

In Table \ref{sec:main_results}, we compare various multimodal foundation models on VQA datasets. Our objective is to evaluate the performance of the \data{} model against the task-specific fine-tuning of \llava-Med on diverse VQA datasets independently. The results for close-ended questions from the VQA-RAD, SLAKE, and PathVQA datasets are presented in Table \ref{tab:task_specific_ft_comparison}.

We observe that \data{} outperforms \llava-Med finetuned on the VQA datasets for three epochs, achieving improvements of $8.8\%$ on VQA-RAD, $24.2\%$ on SLAKE, and $2.3\%$ on PathVQA. Furthermore, we note that \data{} performs better than \llava-Med finetuned for 15 epochs on individual datasets for two out of the three VQA datasets (SLAKE and PathVQA).

These results highlight that a single \data{} model checkpoint is not only highly capable but also more practical for users, eliminating the need to maintain separate task-specific model checkpoints for popular VQA datasets.

\section{Model and Finetuning Details}
\label{app:finetuning_details}

Chameleon \cite{team2024chameleon} represents the raw images as discrete visual tokens using a VQGAN \cite{esser2021taming}, and the text data into discrete text tokens using BPE tokenizer \cite{sennrich2015neural}. Subsequently, each instance in the training dataset is represented as a sequence of discrete tokens, and the model is trained to predict the next token based on the preceding tokens in the sequence, following an autoregressive objective. The model consists of a vocabulary size of $65536$ where $8192$ are visual tokens apart from beginning of image and end of image tokens. In addition, the vocabulary includes a reserved token `<reserved08706>' that separates the instruction (context) from the response (output) for instruction-tuning. Post-tokenization, the entire \data{} data consists $1.7$B tokens where $0.7$B and $1$B are visual and text tokens, respectively.

While the Chameleon-7B model weights are publicly available, they were safety-tuned and support mixed-modal inputs and text-only output to be used for research purposes.\footnote{\url{https://ai.meta.com/blog/meta-fair-research-new-releases/}} To unlock the mixed-modal output capabilities, Anole \cite{chern2024anole} selectively finetunes the output embeddings of the image tokens using high-quality images from LAION \cite{laion}. This strategy does not interfere with the input mixed-modal understanding and text-only output abilities of the original Chameleon model.

We fine-tune the Anole \cite{chern2024anole}, an instantiation of Chameleon \cite{team2024chameleon}, on the \data{} dataset, which consists of $1.47$ million instances. Specifically, we employ low-rank adaptation (LoRA) \cite{hu2021lora} for fine-tuning, using $r=16$, $\alpha=16$, and $\text{dropout}=0.05$. The target modules include the $\{$query, key, value, output, up, down, and gate$\}$ projection matrices. In total, this approach updates $40$M parameters during fine-tuning. We train the model for $3$ epochs using a cosine learning rate schedule (peak LR=$1\text{e}\text{-}4$ with a warmup ratio of $0.1$) and a batch size of $8$. The training is conducted on $8$ Nvidia L40S GPUs (46GB GPU VRAM each).

\section{Additional Image Captioning Results}
\label{app:add_image_captioning_results}

We compare the performance of the \data{} model with other relevant captioning models on the image captioning task using BioCLIPMedScore. The results are presented in Table \ref{tab:add_image_captioning_results}. We observe that the average BioMedCLIPScore for our model is better \llava-Med-v1.5 and at par with baselines such as HuatuoGPT-Vision, GPT-4o-mini, and GPT-4o. Overall, this underscores the capability of the \data{} model in training robust mixed-modal models that excel in diverse biomedical tasks.

\begin{table*}[h]
\centering
\caption{\textbf{Comparison between \data{} model and other baselines on the biomedical image captioning using BioMedCLIPScore.}}
\label{tab:add_image_captioning_results}
\begin{tabular}{lcccc}
\toprule
               & \textbf{Average} & \textbf{PMC}  & \textbf{Quilt} & \textbf{MIMIC-CXR} \\\hline
LLaVA-Med-v1.5 \cite{li2024llavamed} & 0.22    & 0.25 & 0.25  & 0.17      \\
MedMax (Ours)         & 0.38    & 0.37 & 0.36  & 0.41      \\
HuatuoGPT-Vision \cite{zhang2023huatuogpt}        & 0.38    & 0.38 & 0.36  & 0.40       \\
GPT-4o \cite{hurst2024gpt}         & 0.38    & 0.40  & 0.37  & 0.38     \\
\bottomrule
\end{tabular}
\end{table*}

\section{Qualitative Examples}
\label{app:qualitative_examples}

\begin{figure*}[h]
    \centering
        \includegraphics[width=\textwidth]{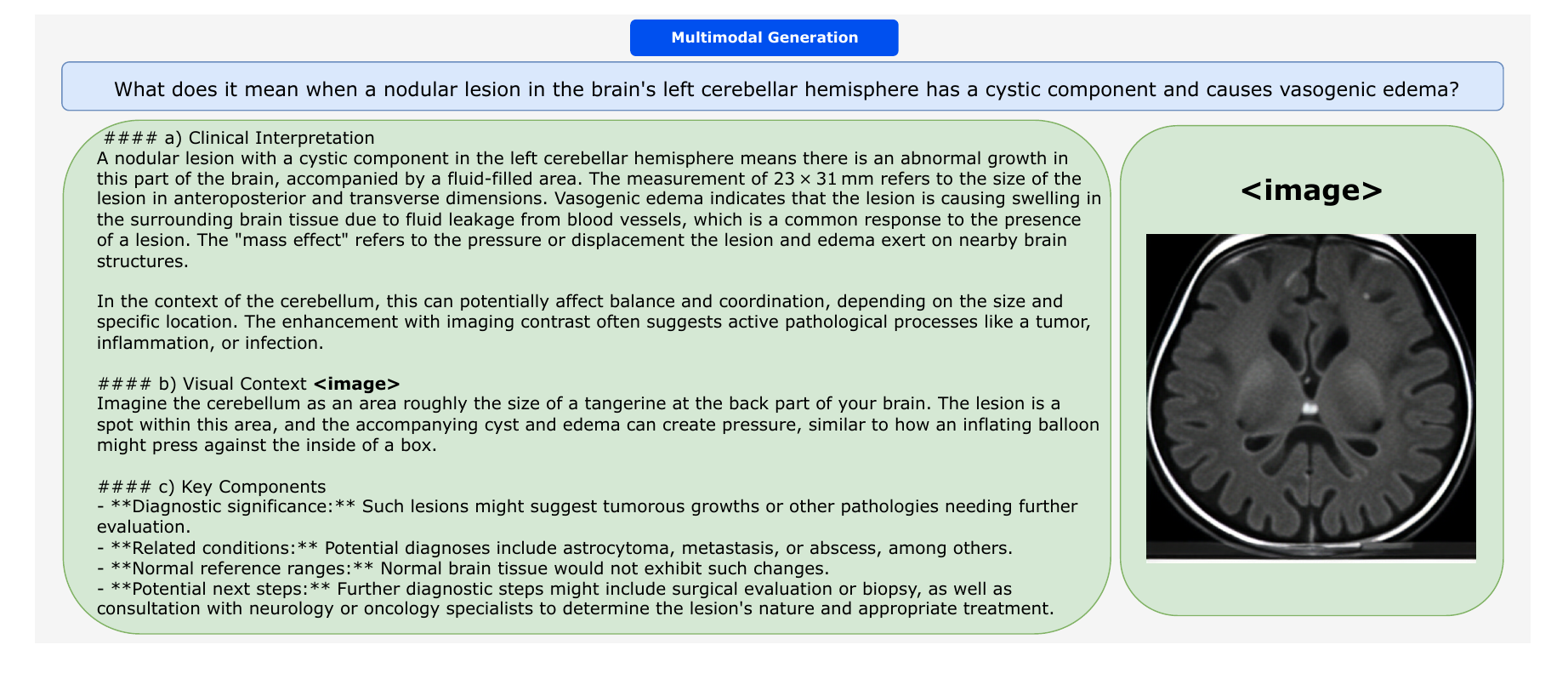}
        \includegraphics[width=\textwidth]{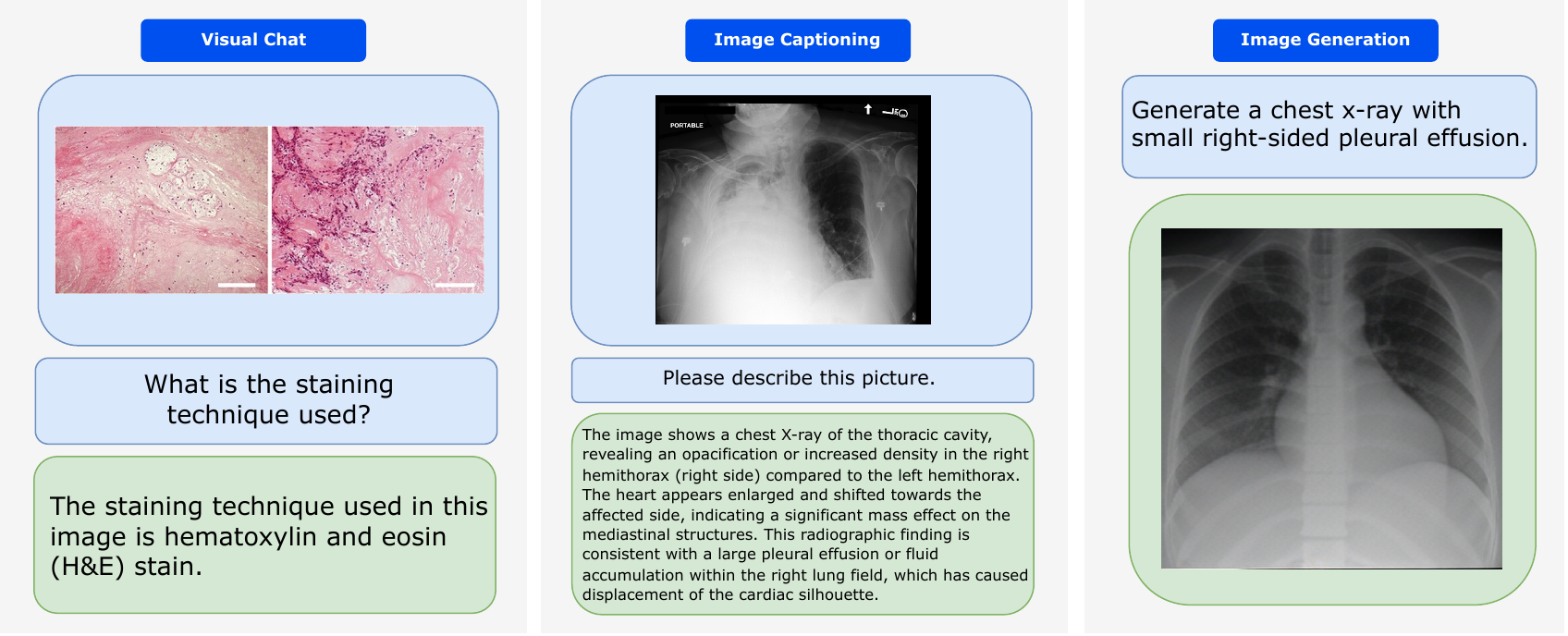}
\caption{\small{\textbf{Positive qualitative examples.} Sample generations from the \data{} model across diverse tasks: multimodal (interleaved) generation, visual chat, image captioning, and image generation. We verify with experts cited in acknowledgments that the provided medical information is high quality and correct.}}
\label{fig:positive_qualitative_examples}
\end{figure*}

\begin{figure*}[h]
    \centering
        \includegraphics[width=\textwidth]{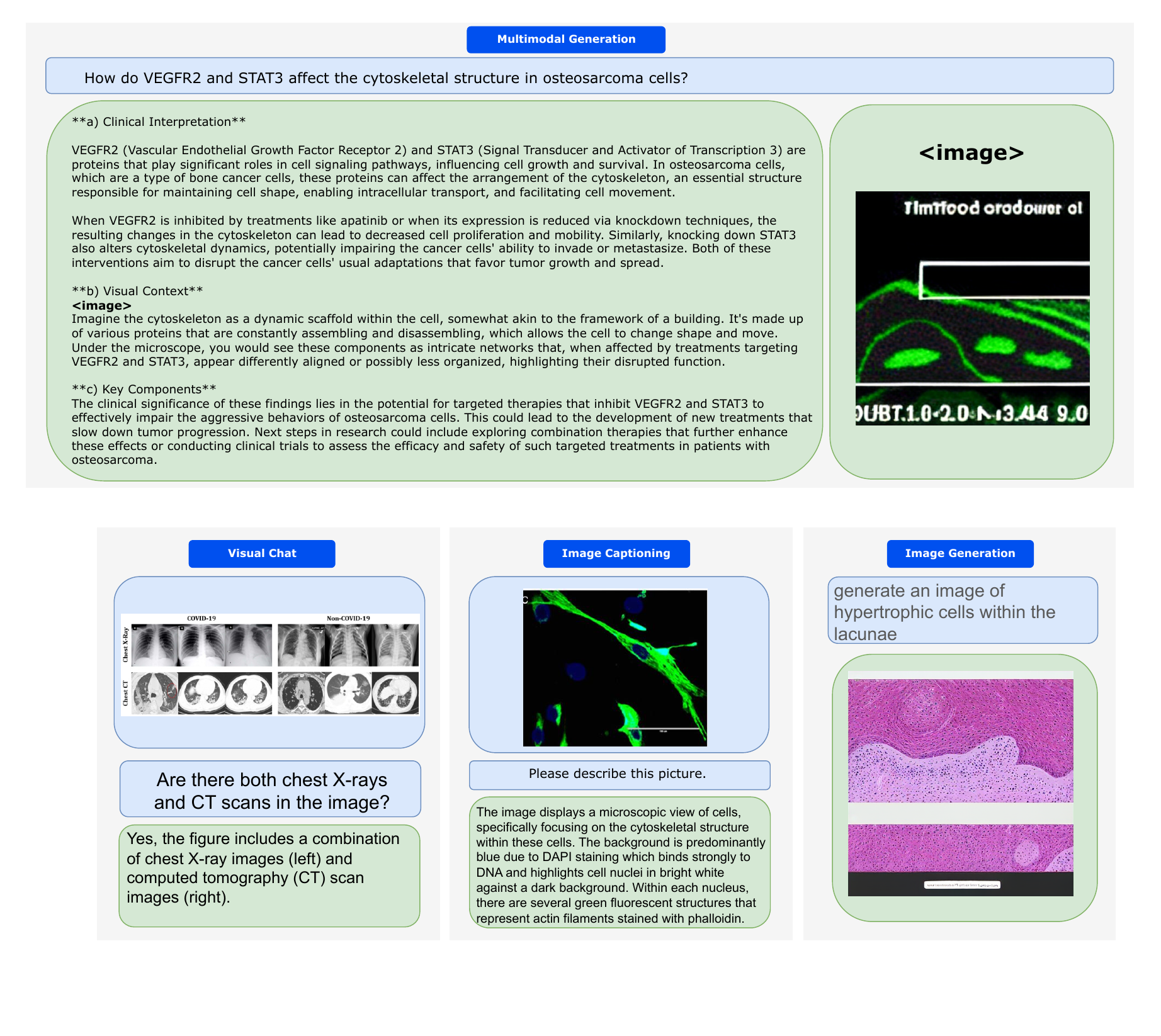}
\caption{\small{\textbf{Negative qualitative examples.} Sample generations from the \data{} model across diverse tasks: multimodal (interleaved) generation, visual chat, image captioning, and image generation. Pitfalls include poor image generation, confusion between segments within an image, and misaligned captions.}}
\label{fig:negative_qualitative_examples}
\end{figure*}

We present the qualitative examples in Figure \ref{fig:positive_qualitative_examples} and Figure \ref{fig:negative_qualitative_examples}.

\end{document}